\documentclass[10pt,letterpaper]{article}
\usepackage[top=0.85in,left=1.75in,footskip=0.75in]{geometry}

\usepackage{changepage}

\usepackage[utf8]{inputenc}

\usepackage{textcomp,marvosym}

\usepackage{fixltx2e}

\usepackage{amsmath,amssymb}

\usepackage{cite}

\usepackage{nameref,hyperref}

\usepackage{microtype}
\DisableLigatures[f]{encoding = *, family = * }

\usepackage{rotating}

\raggedright
\usepackage[aboveskip=1pt,labelfont=bf,labelsep=period,justification=raggedright,singlelinecheck=off]{caption}

\makeatletter
\renewcommand{\@biblabel}[1]{\quad#1.}
\makeatother

\date{}

\usepackage{subfig}
\usepackage[ruled]{algorithm2e}
\usepackage{color}
\usepackage{hyperref}

\usepackage{amsthm}
\theoremstyle{plain}
\newtheorem{theorem}{Theorem}

\usepackage{algpseudocode}

\newcommand{\xv}[0]{\ensuremath{\textbf{x}}\xspace}
\newcommand{\nv}[0]{\ensuremath{\textbf{n}}\xspace}
\newcommand{\ov}[0]{\ensuremath{\textbf{o}}\xspace}
\newcommand{\gv}[0]{\ensuremath{\textbf{g}}\xspace}
\newcommand{\beq}{\begin{equation}}
\newcommand{\eeq}{\end{equation}}
\newcommand{\beqs}{\begin{eqnarray}}
\newcommand{\eeqs}{\end{eqnarray}}

\begin{document}
\vspace*{0.35in}

\begin{flushleft}
{\Large
\textbf\newline{Neuroprosthetic decoder training as imitation learning}
}
\newline
\\
Josh Merel\textsuperscript{1,3,*},
David Carlson\textsuperscript{2,4},
Liam Paninski\textsuperscript{1,2,3,4},
John P. Cunningham\textsuperscript{1,2,3,4}
\\
\bigskip
\bf{1} Neurobiology and Behavior program, Columbia University, New York, NY, USA
\\
\bf{2} Department of Statistics, Columbia University, New York, NY, USA
\\
\bf{3} Center for Theoretical Neuroscience, Columbia University, New York, NY, USA
\\
\bf{4} Grossman Center for the Statistics of Mind, Columbia University, New York, NY, USA
\bigskip

%
%





* jsmerel@gmail.com

\end{flushleft}

\section*{Abstract}
Neuroprosthetic brain-computer interfaces function via an algorithm which decodes neural activity of the user into movements of an end effector, such as a cursor or robotic arm. In practice, the decoder is often learned by updating its parameters while the user performs a task.  
When the user's intention is not directly observable, recent methods have demonstrated value in training the decoder against a surrogate for the user's intended movement.
We describe how training a decoder in this way {is a novel variant of an} \textit{imitation learning} problem, where an oracle or expert is employed for supervised training in lieu of direct observations, which are not available.     
Specifically, we {describe how} a generic imitation learning meta-algorithm, dataset aggregation (\textsc{DAgger}, \cite{Ross2011}), can be {adapted} to train a {generic} brain-computer interface.  
By {deriving} existing learning algorithms for brain-computer interfaces in this framework, we provide { a novel analysis of \textit{regret} (an important metric of learning efficacy) for brain-computer interfaces}.
{This analysis allows us} to characterize the space of algorithmic variants and bounds on their regret rates.
Existing approaches for decoder learning have been performed in the cursor control setting, but 
{the available design principles for these decoders are such that it has been impossible to scale them to naturalistic settings.}
{Leveraging our findings, we then offer an algorithm that combines imitation learning with optimal control, which should allow for training of arbitrary effectors for which optimal control can generate goal-oriented control.  We demonstrate this novel and general BCI algorithm with simulated neuroprosthetic control of a 26 degree-of-freedom model of an arm, a sophisticated and realistic end effector.}

\section*{Author Summary}
There are various existing methods for rapidly learning a decoder during closed-loop brain computer interface (BCI) tasks. While many of these methods work well in practice, there is no clear theoretical foundation for parameter learning.
We offer a {unification} of closed-loop decoder learning setting as an imitation learning problem. 
This has two major consequences: first, our approach clarifies how to {derive} “intention-based” algorithms for any BCI setting, most notably more complex settings like control of an arm; and second, this framework allows us to {provide} theoretical results, {building from} an existing literature on the regret of related algorithms.  
After first demonstrating algorithmic performance in simulation on the well-studied setting of a user trying to reach targets by controlling a cursor on a screen, we then simulate a user controlling an arm in order to grasp a wand.  Finally, we {describe how extensions in the online-imitation learning literature can improve BCI in additional settings.}

\section*{Introduction}

Brain-computer interfaces (BCI, or brain-machine interfaces) translate noisy neural activity into commands for controlling an effector via a decoding algorithm \cite{serruya2002brain,taylor2002direct,carmena2003learning,hochberg2006neuronal}. 
{ 
While there are various proposed and debated encoding mechanisms describing how motor-relevant variables actually relate to neural activity \cite{georgopoulos1984static,moran1999motor,todorov2000direct,georgopoulos2000one,churchland2012neural}, in practice decoders are successful at leveraging the statistical relationship between the intended movements of the user and firing rates of recorded neural signals.
Under the operational assumption that some key variables of interest (e.g. effector kinematics) are linearly encoded by neural activity, the Kalman filter (KF) is a reasonable decoding approach \cite{wu2006bayesian}, and empirically it yields state-of-the-art decoding performance \cite{gilja2012high} (see \cite{zhang2015recasting} for review).  
Once a decoder family (e.g. KF) is specified, a core objective in decoder design is to obtain good performance by learning specific parameter values during a training phase.  
For a healthy user who is capable of making overt movements (as in a laboratory setup with non-human primates \cite{serruya2002brain,taylor2002direct,carmena2003learning,gilja2012high}), it is possible to observe neural activity and overt movements simultaneously in order to directly learn the statistical mapping -- implicitly, we assume the overt movements reflect \textit{intention}, so this mapping provides a relationship between neural activity and intended movement.  
}

However, in many cases of interest the user is not able to make overt movements, so intended movements must be inferred or otherwise determined.  This insight that better decoder parameters can be learned by training against some form of assumed intention appears in \cite{gilja2012high}, and extensions have been explored in \cite{dangi2013design,dangi2014continuous}.
In these works, it is assumed that the user intends to move towards the current goal or target in a cursor task, resulting in parameter training algorithms that result in dramatically improved decoder performance on a cursor task.

Specifically, in the \textit{recalibrated feedback intention-trained Kalman filter} formulation (ReFIT, \cite{gilja2012high}), the decoder is trained in two stages. 
First, the subject makes some number of reaches using its real arm.  
The hand kinematics and neural data are used to train a Kalman filter decoder.  
Next, the subject engages in the reach-task in an online setting using the fixed Kalman filter decoder.  The decoder could be updated naively with the data from this second stage (gathered via closed loop control of the cursor).  
However, the key parameter-fitting insight of ReFIT is that a demonstrably better decoder is learned by first modifying this closed-loop data to reflect the assumption that the user intended at each timestep to move towards the target (rather than the movement that the decoder actually produced).
Specifically, the modification is that the instantaneous velocity from the closed-loop cursor control is \textit{rotated} to point towards the goal to create a goal-oriented dataset.  
The decoder is then trained on this modified dataset.  
ReFIT additionally proposes a modified decoding algorithm.
However, we emphasize the distinction between the problem of learning parameters and selection of the decoding algorithm -- this paper focuses on the problem of learning parameters (for discussion concerning decoding algorithm selection, see \cite{zhang2015recasting}).

Shortcomings of ReFIT include both a lack of understanding the conditions necessary for successful application of its parameter-fitting innovation, as well as the inability for the user to perform overt movements required for the initial data collection when the user is paralyzed (as would be the norm for clinical settings \cite{hochberg2006neuronal,hochberg2012reach,gilja2015clinical}).  
{
But even more critical an issue is that ReFIT is exclusively suited to the cursor setting by requiring the intuitively-defined, goal-rotated velocities. 
The \textit{closed-loop decoder adaptation} (CLDA) framework has made steps towards generalizing the ReFIT parameter-fitting innovation \cite{dangi2013design}.
The CLDA approach built on ReFIT, effectively proposing to update the decoder online as new data streamed in using an adaptive scheme \cite{dangi2013design,dangi2014continuous}.  While these developments significantly improve the range of applicability, they still rely on rotated velocities and do not address the key issue of extending these insights to more complex tasks, such as control with a realistic multi-joint arm effector.
In the present work, we provide a clear approach which generalizes this problem to arbitrary effectors and contextualizes the style of parameter fitting employed in both ReFIT and CLDA approaches as special cases of a more general online learning problem, called ``imitation learning."
}

In imitation learning (or ``apprenticeship learning"), an agent must learn what action to take when in a particular situation (or state) via access to an expert or oracle which provides the agent with a good action at each timestep.  
The agent can thereby gradually learn a policy for determining which action to select in various settings.  This setting is related to online learning \cite{Ross2011}, wherein an agent makes sequential actions and receives feedback from the environment regarding the quality of the action.  
{
We propose that, in the BCI setting, instead of a policy that asserts which action to take in a given state, we have a decoder that determines the effector update in response to the current kinematic state and neural activity.  
Formally, the decoder serves the role of the policy; the neural activity and the current kinematic pose of the effector comprise the state; and the incremental updates to the effector pose correspond to actions.
We also formalize knowledge of the user's instantaneous ``true" intention as an \textit{intention-oracle}.  With this oracle, we can train the decoder in an online-imitation data collection process using update rules that follow from supervised learning. 
}

{

Our work helps to resolve core issues in the application of intention-based parameter fitting methods.   
(1) By explicitly deriving intention-based parameter fitting from an imitation learning perspective, we can describe a family of algorithms, provide general guarantees for the closed-loop training process, and provide specific guarantees for standard choices of parameter update rules.  
(2) We generalize intention-based parameter fitting to more general effectors through the use of an optimal control solver to generate an intention-oracle.  We provide a concrete approach to derive goal-directed intention signals for a model monkey arm in a reaching task. Simulations of the arm movement task demonstrate the feasibility of leveraging intention-based parameter fitting in higher dimensional tasks -- something fundamentally ambiguous given existing work, because it was not possible to infer intention for high-dimensional tasks or arbitrary effector DOF representations.  
}

In the next section, we formulate the learning problem.
We then present a family of CLDA-like algorithms which encompasses existing approaches. By relating BCI learning algorithms to their general online learning counterparts in this way, we can leverage the results from the larger online learning literature.
We theoretically characterize the algorithms in terms of bounds on ``regret." Regret is a measure of the performance of a learning algorithm relative to the performance if that algorithm were set to its optimal parameters.  
However, while bounds are highly informative about dominant terms, they are often ambiguous up to proportionality constants.  Therefore,	we employ simulations to give a concrete sense of how well these algorithms can perform and provide a demonstration that even learning to control a full arm is now feasible using this approach.

\section*{Results}

\subsection*{Components of the imitation learning approach for BCI}

\begin{figure}[t!]
\begin{center}
  \subfloat{\includegraphics[width=.7\textwidth]{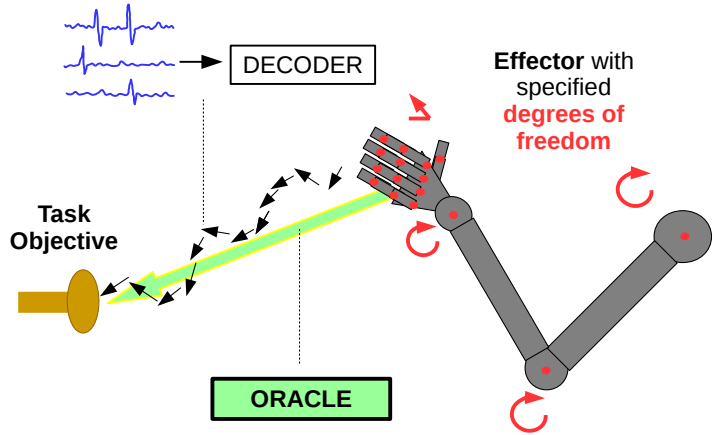}} 
  \end{center}    
  \captionsetup{width=.8\textwidth}
\captionsetup{font={footnotesize}}
\caption{A BCI has an \textit{effector}, such as a robotic arm, with predefined degrees of freedom.  Given a \textit{task objective} (e.g. an objective function corresponding to reaching and grasping a target), an \textit{intention-oracle} can be computed to provide the intended updates to the arm kinematics.  The actual trajectory of the arm is evaluated deterministically from the neural activity via the \textit{decoder}.  In practice, the oracle update would be recomputed at each timestep to reflect the instantaneous best movement in the direction of the goal.}
\label{fig:diagram_1}
\end{figure}

The problem that arises in BCI parameter fitting is to learn the parameters of the model in an online fashion.  In an ideal world, this could be performed by supervised learning, where we observe both the neural activity and overt movements, which reflect user intention.  In a closed-loop setting, we would then simply use supervised online learning methods.  However, for supervised learning we need labelled movement data.  Neither overt movements nor user intent are actually observable in a real-world prosthetic setting.  Imitation learning, through the usage of an oracle or expert, helps us circumvent this issue.
To begin, we describe the core components of BCI algorithms that follow the imitation learning paradigm -- effector, task objective, oracle, decoding algorithm, and update rule (Fig. \ref{fig:diagram_1}).  

The \textit{effector} for a BCI is the part of the system that is controlled in order to interact with the environment (e.g. a cursor on a computer screen \cite{gilja2012high} or a robotic arm \cite{hochberg2012reach,Putrino201568}).  
Minimally, the degrees of freedom (DOF) that are able to be controlled must be selected.
For example, when controlling a robotic arm, it might be decided that the user only controls the hand position of the robotic arm (e.g. as if it were a cursor in 3D) and the updates to the arm joint angles are computed by the algorithm to accommodate that movement.  
A model of effector dynamics provides a probabilistic state transition model, which permits the use of filtering techniques as the decoding algorithm.  The default assumption for dynamics is that the effector does not move discontinuously, which yields smoothed trajectories. 

The \textit{task objective} refers to the performance measure of the task.  For example, in a cursor task, the objective could be for the cursor to be as close as possible to the goal as rapidly as possible, or it may be for the cursor to acquire as many targets as possible in some time interval.  Other objectives related to holding the cursor at the target with a required amount of stability have also been proposed (e.g. ``dial-in-time" as in \cite{gilja2012high}).
The objective may include be additional components related to minimizing exertion (i.e. energy) or having smooth/naturalistic movements.
Insofar as this task has been communicated to the user (verbally in the human case or via training in the case of non-human subjects), the user's intention should be consistent with this objective, so it is appropriate to consider the task objective to correspond to the user's intended objective.  

Imitation learning requires an \textit{oracle} or expert to provide the labelled data.  When overt movements are available, we use overt movements as a proxy for the intended movements.  { Retrospectively, we re-interpret the parameter-fitting innovation of ReFIT in the imitation learning framework -- specifically, the choice to train using goal-directed velocity vectors \cite{gilja2012high} was an implicit selection of intention-oracle (a model of the user's intention).}  Indeed this is a reasonable choice of oracle as it is goal-directed, presumably reflects user intent, and provides a sensible heuristic for the magnitude of the instantaneous oracle velocities.  More generally, the oracle should be selected to match the user’s intention as closely as possible (for example by compensating for sensory delays as in \cite{golub2013learning}).  When the task objective is well-specified and there exists a dynamics model for the effector, routine optimal control theory can be used to produce the oracle (along the lines of \cite{Ross12agnosticsystem}).  That is, from the current position, the incremental update to the effector state in the direction of the task objective can be computed.  For a cursor, a simple mean-squared error (MSE) objective will result in optimal velocities directed towards the goal/target, with extra assumptions governing the magnitudes of those velocities.

Different BCI algorithms also differ in their choice of \textit{decoder} family and \textit{update rule}.  
{We can abstract these decoders as learned functions} mapping neural activity and current effector state to kinematic updates (e.g. this is straightforward for the steady-state Kalman filter, see methods).  The parameters of the model will be adapted by an update rule, which makes use of the observed pairs of data (i.e the intention-oracle and the neural activity).
We note two complementary perspectives -- we can use our data to directly update decoder parameters or alternatively we can update the encoding model parameters and compute the corresponding updated optimal decoder (i.e. using Bayes rule to combine the encoding model and the effector dynamics model to decode via Bayesian filtering).  
In principle, either of these approaches work, but in this work we will directly adapt decoder parameters because it is simpler and closer to the convention in online learning.      

{

In very general decision process settings, a function mapping from states to actions is called a \textit{policy} \cite{Bel,lavalle2006planning} -- in BCI settings, this is the decoder. The details of this mapping can be specified in a few essentially equivalent ways.  Most consistent with the state-action mapping is for the policy to produce an action corresponding to an update to the state of the effector.  If the effector state consists of positions, then these updates are velocities; but the effector state could also be instantaneous velocities, forces, or other variables, in which case the actions correspond to updates to these state variables and imply updates to the pose of the effector. 

Relatively more familiar in BCI research is the use of a policy as decoder when reinforcement learning (RL) is being used (see \cite{mahmoudi2013towards,bryan2013probabilistic,pohlmeyer2014using}, or even with error feedback derived from neural activity in other brain regions \cite{iturrate2015teaching}).  Reinforcement learning and imitation learning involve similar formalisms.  However, the most suitable learning framework depends on the available information.  
Conventional RL only provides information when feedback is available (e.g. when the task is successful), whereas use of an oracle in imitation learning allows for training informed by every state.
This will yield considerably more rapid learning than RL.  
There are various ways to learn a policy using frameworks between these extremes.  In an actor-critic RL framework \cite{sutton1998reinforcement}, the policy (a.k.a. actor) is trained from a learned value function (a.k.a. critic)
 -- readers familiar with this framework may see this as a conceptual bridge between imitation learning and RL, where imitation learning uses oracle examples rather than a learned value function.  
It is also possible to learn an expert’s reward function from examples and directly train the policy \cite{abbeel2004apprenticeship}. 
Perhaps most usefully, a policy could also be learned from hybrid RL and imitation updates, and this would be well-advised if the oracle is noisy or of otherwise low quality (see Discussion).
}
    
\subsection*{Parameter updating through imitation learning}

{
We next present a BCI meta-algorithm which formalizes closed-loop data collection and online parameter updating as a variant of imitation learning.  
This perspective is consistent with the CLDA framework \cite{dangi2013design}, but by formalizing the entire approach as a meta-algorithm, we gain additional theoretical leverage. 
BCI training as described in this meta-algorithm amounts to a non-standard imitation learning setting insofar as the oracle comes from a task-constrained model of user intention, and the decoder is a policy that is conditioned on noisy neural activity.  
The imitation learning formalization of this BCI learning procedure is consistent with the online-imitation learning framework and meta-algorithm \textit{dataset aggregation}  (\textsc{DAgger}) \cite{Ross2011}.
We will subsequently show the online-imitation learning framework encompasses a range of reasonable closed-loop BCI approaches.  
}

We set up the process such that the data is split into reach \textit{trajectories} $k=1,\dots,K$ that each contain a sequence of $T_k < T$ discretized time points, and $K$ is not necessarily known \textit{a priori}.
Each $T_k$ corresponds to the time it takes for a single successful reach.
The $k^{th}$ reach is successful when some task objective, such as the distance between the cursor position and a goal position $\gv_{kt}$, is satisfied to within some $\epsilon$ (more generally, the goal $\gv_{kt}$ corresponds to any sort of target upon which the objective depends).  
At each time point within a reach, $t$, we assume that we have the current state of the effector $\xv_{kt}$, as well as a vector $\nv_{kt}$ that corresponds to neural activity (e.g. spike counts).  Bold lower-case letters ($\xv,~\nv,~\gv,\dots$) denote column vectors.
{
The decoder will update the state of the effector based on the combined neural state and previous effector state, $\{\nv_{kt},\xv_{kt}\}$ 
(in a limiting case, the decoder may only rely on neural activity, but inclusion of previous effector states allows for smoothing of effector trajectories).

Formally, }we want a decoder $\pi \in \Pi$ (i.e., a \textit{policy} $\pi$ within the space of policies $\Pi$) that transforms the state information ($\xv,\nv$) into an action that matches the intention of the user.
An imitation learning algorithm trains the policy to mimic as closely as possible the oracle policy $\pi^*$, which gives the oracle actions $\ov_{kt} = \pi^*(\xv_{kt},\nv_{kt},\gv_{kt})$.  Note that the oracle policy is not a member of $\Pi$ (i.e. $\pi^* \notin \Pi$): this distinction is important as the learnable policies $\pi$ do not have access to goal information.
Because we have finite samples, we use an instantaneous loss $\ell(\pi(\xv_{kt},\nv_{kt}),\ov_{kt})$. In the cursor control case, this loss could be the squared error between the oracle velocity and the decoder/policy velocity.
We write $\mathcal{L}(\pi, \mathcal{D}^{(1:k)})$ as shorthand for $\sum_{k=1}^K \sum_{t=1}^{T_k} \ell(\pi(\xv_{kt},\nv_{kt}),\ov_{kt})$, where $\mathcal{D}^{(k)}$ refers to the set of data $\{\xv_{kt},\nv_{kt},\ov_{kt}\}_{t \in 1...T_k}$ from just the $k^{th}$ reach, and $\mathcal{D}^{(1:k)}$ refers to the combined set of data $\{\xv_{k^{\prime} t},\nv_{k^{\prime} t},\ov_{k^{\prime} t}\}_{k^{\prime} \in 1...k, t \in 1...T_{k^{\prime}}}$ from reaches up to $k$.

\begin{algorithm}[]
	
    \caption{Imitation learning perspective of decoder training \label{alg_main}} 
	Initialize dataset $ \mathcal{D}^{(0)} \leftarrow \emptyset$ \\
	Initialize decoder $\pi^{(0)}$ \\
	Input/select $\beta_1,\dots,\beta_K$
    \BlankLine
    \For{$k = 1$ \KwTo $K$ trajectories}{
    		Initialize effector state, $\xv_{k1} \leftarrow \xv_0$, (or continue from end of previous trajectory)\\
    		Randomly select goal state, $\gv_{kt}$ from set of valid goals\\
    		Initialize $t \leftarrow 1$\\
    		\While{$distance($\emph{$\xv_{kt},\gv_{kt}$}$)>\epsilon$ \textbf{and} $t < T$}{
    			Acquire neural data $\nv_{kt}$\\
    			Query oracle update $\ov_{kt}=\pi^*(\xv_{kt},\nv_{kt},\gv_{kt})$\\
			Update state via assisted decoder: \\
			~~~~~$\xv_{k,t+1} \leftarrow \beta_k\pi^*(\xv_{kt},\nv_{kt},\gv_{kt})+(1-\beta_k)\pi^{(k)}(\xv_{kt},\nv_{kt})$\\
    			$t \leftarrow t+1$
    		} 
		Aggregate $ \mathcal{D}^{(1:k+1)} \leftarrow \mathcal{D}^{(1:k)} \cup \left\{(\xv_{kt},\nv_{kt},\ov_{kt})\right\}_{t=1,\dots,T_k}$\\
		$\pi^{(k+1)} \leftarrow \textsc{{update}}(\pi^{(k)},\mathcal{D}^{(1:k+1)})$ (See Alg. \ref{dec_updates})
    }
    \Return{best or last $\pi$}
\end{algorithm} 

The core imitation learning \textit{meta}-algorithm is presented in Alg. \ref{alg_main}.  
This meta-algorithm describes the general structure for different learning algorithms, and the \textsc{{update}} line is distinct for alternative learning methods { (each \textsc{{update}} takes the current decoder and dataset and produces the new decoder).} 
We emphasize that this meta-algorithm is specified only once the effector, task objective, oracle, decoding algorithm, and parameter update rule are determined.
The \textsc{DAgger} process gradually aggregates a dataset $\mathcal{D}$ with pairs of state information and oracle actions at each time point.
The dataset is used to train a stationary, deterministic decoder, which is defined as the deterministic optimal action (lowest average loss) based on the state information, which includes both the neural activity ($\nv$) and the effector state ($\xv$) in the BCI setting.

{The meta-algorithm begins with an initial decoder (i.e. stable, albeit poorly performing) and uses this decoder, possibly blended with the oracle, to explore states.} 
Specifically, the effective decoder 
is given by $\beta_i \pi^* + (1-\beta_i)\pi^{(k)}$, where $\pi^*$ is the oracle policy and $\pi^{(k)}$ is the current decoder.  
{ When this mixing is interpreted as a weighted linear sum, t}his approach is equivalent to \textit{assisted decoding} in the BCI literature (as in \cite{velliste2008cortical} or \cite{So2014}), where the effective decoder during training is a mixture of the oracle policy and the decoder driven by the neural activity { -- in \cite{Ross2011}, the policy blending is probabilistic (see Text S1 for detailed distinction)}.  
The assisted decoder may reduce user frustration from poor initial decoding, and helps provide more task-relevant sampling of states.  
As training proceeds, the effective decoder relies less on the oracle and is ultimately governed only by the decoder.
For example, $\beta_i$ may be set to decrease { according to a particular schedule with iterations, or as an abrupt example,} $\beta_1 = 1$ and $\beta_{i>1} = 0$.  

For each time point in each trajectory, the state information and oracle pair are incorporated into the stored dataset.  
The decoder is updated by a chosen rule at the end of each trajectory (or alternatively after each time step).
We note that computational and memory requirements are less for updates that only require data from the most recent stage ($\mathcal{D}^{(k+1)}$);
however, using the whole dataset is more general, may improve performance, and can stabilize updates.

\subsection*{Relating BCI and online learning}

{
Imitation learning with an intention-oracle is a natural framework to re-interpret and understand the parameter fitting insights that were proposed in the ReFIT algorithm \cite{gilja2012high}.
In the ReFIT work, the authors used modified velocity vectors in order to update parameters in a fashion which incorporated the user's presumed goal-directed intention, and this approach was empirically justified.  We can re-interpret the rotated vectors as an \textit{ad hoc} oracle, with these vectors and the single batch re-update being specific choices, hand-tailored for the task.
}

The CLDA framework extracted the core parameter-fitting principle from ReFIT, allowing for the updates to occur multiple times and take different forms \cite{dangi2013design}.  The simplest update consistent with this framework is \textit{gradient-based decoder adaptation}. Under this scheme the decoder is repeatedly updated and the updates correspond to \textit{online gradient descent} (OGD).  
This general class of BCI algorithms take observations in an online fashion, perform updates to the parameters using the gradient, and do not pass over the ``old" data again.  This \textsc{{update}} takes the form:
\beq
\label{OGD_eq}
\pi^{(k+1)} = \pi^{(k)} - \frac{1}{\eta_k} \nabla_{\pi} \mathcal{L}(\pi^{(k)},\mathcal{D}^{(k+1)}),
\eeq
which simply means that decoder parameters are updated by taking a step in the direction of the negative gradient of the loss with respect to those parameters.  $\frac{1}{\eta_k}$ corresponds to the learning rate.  

{A second option for parameter updating} is to smoothly average previous parameter estimates with recent (temporally localized) estimates of those parameters computed from a mini-batch -- that is, to perform a \textit{moving average} (MA) over recent optimal parameters.  This \textsc{{update}} takes the form: 
\beq
\label{MA_eq}
\pi^{(k+1)} = (1-\lambda)\pi^{(k)} + \lambda \arg \min_{\pi} \mathcal{L}(\pi,\mathcal{D}^{(k+1)})
\eeq
for $\lambda\in[0,1]$.  In practice, the second term here corresponds to maximum likelihood estimation of the parameters.  An update of this sort is presented as part of the CLDA framework as \textit{smoothBatch} \cite{dangi2013design}.  

{A third parameter update option} in the BCI setting is to peform a full re-estimation of the parameters given all of the observed data at every update stage.  This can be interpreted as a \textit{follow-the-leader} (FTL) update \cite{Shalev-Shwartz2011}.  This \textsc{{update}} takes the form: 
\beq
\label{FTL_eq}
\pi^{(k+1)} = \arg \min_{\pi} \mathcal{L}(\pi,\mathcal{D}^{(1:k+1)}).
\eeq
Here all data pairs are used as part of the training of the next set of parameters.  We will show in the next section that this update can provide especially good guarantees on performance.
\textsc{DAgger} was originally presented using this FTL update, utilizing the aggregated dataset \cite{Ross2011}.  
We note that this sort of batch maximum likelihood update is discussed as a CLDA option in \cite{dangi2014continuous}, where a computationally simpler, exponentially weighted variant is explored, termed \textit{recursive maximum likelihood} (RML).  
For BCI settings, data is costly relative to the memory requirements, so it makes sense to aggregate the whole dataset without discarding old samples.  For all of these updates, especially early on, it can be useful to include regularization, and we also incorporate this into the definition of the loss. 
We summarize the parameter update procedures in Alg. \ref{dec_updates}.  

\begin{algorithm}[t]
    \caption{Selected direct decoder \textsc{{update}} options \label{dec_updates}}
    \textbf{Switch:}\\
    \hspace{1em}\textbf{Case -- Online gradient descent (OGD), Eqn. \ref{OGD_eq}:} \\
	\hspace{2em}$\pi^{(k+1)} = \pi^{(k)} - \frac{1}{\eta_k} \nabla_{\pi} \mathcal{L}(\pi^{(k)},\mathcal{D}^{(k+1)})$ \\
	\hspace{1em}\textbf{Case -- Moving average (MA), Eqn. \ref{MA_eq}:} \\
	\hspace{2em}$\pi^{(k+1)} = (1-\lambda)\pi^{(k)} + \lambda \arg \min_{\pi} \mathcal{L}(\pi,\mathcal{D}^{(k+1)})$ \\
	\hspace{1em}\textbf{Case -- Follow the (regularized) leader (FTL), Eqn. \ref{FTL_eq}: } \\
	\hspace{2em}$\pi^{(k+1)} = \arg \min_{\pi} \mathcal{L}(\pi,\mathcal{D}^{(1:k+1)}) $ \\
	\Return $\pi^{(k+1)} $\\
\end{algorithm} 

{

Adaptive filtering techniques in engineering are closely related to the online machine learning updates we consider in this work.  OGD is a generic update rule.  In the special case of linear models with a mean square error cost, the solution that has a long history in engineering is called the least mean square (LMS) algorithm \cite{Widrow1985}.  Also, in the same setting, when FTL corresponds to a batch LS optimization, its solution could be computed exactly in an online fashion using recursive least squares (RLS) \cite{plackett1950some} (for more background on LMS or RLS see \cite{sayed2003fundamentals}).
}
	
We will more concretely discuss the guarantees of these algorithms in the subsequent section.  We remark that all of the algorithms described so far make use of our generalization of the key parameter-fitting innovation from ReFIT, but they differ in parameter update rule.  Additionally, algorithms can differ in the selection of the decoding algorithm, effector, task objective, and oracle.  For example, if some objective other than mean squared error (MSE) were prioritized (e.g. rapid cursor stopping) and it was believed that user intention should reflect this priority, then the task objective and oracle could be designed accordingly.

\subsection*{Algorithm regret bounds}

In this section we provide theoretical guarantees for the BCI learning algorithms introduced above.  
{Our formalization of the BCI setting allows us to provide new theory for closed-loop BCI learning by combining core theory for \textsc{DAgger} \cite{Ross2011} with adaptations of results from the online learning literature.}  
We provide specific terms and rates for the representative choices of parameter update rules (discussed in previous sections, summarized in Table \ref{table_regret}).  

The standard way of assessing the quality of an online learning algorithm is through a \textit{regret} bound \cite{Shalev-Shwartz2011}, which calculates the excess loss after $K$ trajectories relative to having used an optimal, static decoder from the set of possible decoders ${\Pi}$:
\begin{equation}
\text{Regret}_{K}(\Pi)=\max_{\pi^\flat\in\Pi} \sum_{k=1}^K\sum_{t=1}^{T_k} \left(\ell(\pi^{(k)}(\xv_{kt},\nv_{kt}),\ov_{kt}) -\ell(\pi^\flat(\xv_{kt},\nv_{kt}),\ov_{kt}) \right).\label{regret}
\end{equation}

A smaller regret bound or a regret bound that decays more quickly is indicative of an algorithm with better worst-case performance.
Note that $\pi^\flat$ is the best realizable decoder (${\Pi}$ is the set of feasible decoders, which may have a specific parameterization and will not depend on the goal), so $\pi^\flat$ is not equivalent to the oracle.  Since $\pi^\flat$ will need to make use of noisy neural activity, the term $\ell(\pi^\flat(\xv_{kt},\nv_{kt}),\ov_{kt})$ is not likely to be zero.

{Because we have been able to formulate closed-loop BCI learning as imitation learning, we inherit a variant of the} core theorem of \cite{Ross2011} (see Text S1 for our restatement), which can be paraphrased as stating: \textit{Alg. \ref{alg_main} will result in a policy (i.e. decoder) that has an expected total loss bounded by the sum of three terms: (1) a term corresponding to the loss if the best obtainable decoder had been used for the whole duration; (2) a term that compensates for the assisted training terms ($\beta_k$); (3) a term that corresponds to the regret of the online learning parameter update rule used.}

We emphasize that the power of this theorem is that it allows analysis of imitation learning through regret bounds for well-established online optimization methods. Regret that accumulates sublinearly with respect to observations implies that the trial-averaged loss can be expected to converge.  We usually want the regret accumulation to occur as slowly as possible.  A goal of online learning is to provide \emph{no-regret} algorithms, which refers to the property that $\lim_{K\rightarrow \infty} \text{Regret}_K(\Pi)/K =0$.

\begin{table}[tp]
\caption{Summary of regret for selected algorithms.}
\begin{center}
\begin{tabular}{|c|c|c|}
\hline
 Online Learning Algorithm & Closest BCI Algorithm & Regret \\\hline\hline
Online Gradient Descent & Gradient-based decoder adaptation & $\mathcal{O}(\sqrt{K})$ \cite{Kivinen1995} \\
~&~&$\mathcal{O}(\log{K})^\ast$ \cite{Hazan2007}\\\hline
Moving Average & SmoothBatch \cite{dangi2013design} & $\mathcal{O}(K)$ \\\hline
Follow-the-leader & CLDA-style maximum likelihood \cite{dangi2014continuous} & $\mathcal{O}(\log{K})$ \cite{Hazan2007} \\\hline
\end{tabular}
  \captionsetup{width=.8\textwidth}
\captionsetup{font={footnotesize}}
\caption*{$^{\ast}$ Bound obtained only under restrictive conditions (see main text).}
\end{center}
\label{table_regret}
\end{table}

In this work, we have introduced three update methods that serve as a representative survey of the simple, intuitive space of algorithms proposed for the BCI setting (see Table \ref{table_regret}).  We provide regret bounds for the imitation learning variants, here specifically assuming linear decoding and a quadratic loss (see Text S1 for full details).  
This analysis is based on the \textit{steady-state Kalman filter}  (SSKF) (see methods), but could be generalized to other settings. 

OGD is a classical online optimization algorithm, and is well-studied both generally and in the linear regression case.  The regret scales as $\mathcal{O}(\sqrt{K})$ \cite{Kivinen1995} (recall $k$ indexes the reach trajectory).  We note that in order to saturate the performance of OGD, the learning rate must be selected carefully, and the optimal learning rate essentially requires knowledge of the scaling of the parameters.  
Each parameter may require a distinct learning rate for optimal performance \cite{duchi2011adaptive}.  OGD is most useful in an environment where data is cheap because the updates have very low computational overhead -- this is relevant for many modern large-data problems.  
In BCI applications, data is costly due to practical limits on collecting data from a single subject, so a more computationally intensive update may be preferable if it outperforms OGD.    

Under certain conditions, OGD can achieve a regret rate of $\mathcal{O}(\log K)$ \cite{Hazan2007}, which is an improved rate (and the same order as the more computationally-intensive FTL strategy we discuss below).  This rate requires additional assumptions that are realistic only for certain practical settings.  Asymptotically, any learning rate $\eta_k$ that scales as $\mathcal{O}(k)$ will achieve this logarithmic rate, but choosing the wrong scale will dramatically negatively impact performance, especially during the crucial, initial learning period.
For this reason, we may desire methods without step-size tuning.

We next provide guarantees available for the \textit{moving average} update.  This algorithm suffers from regret that is $O(K)$, so it is not a no-regret algorithm (see analysis presented in \cite{dangi2013design} where there is an additional steady-state error).  Conceptually this is because old data has decaying weight, so there is estimation error due to prioritization of a recent subset of the data.  
While this method has poor regret when analyzed for a \textit{static model} (i.e. neural tuning is stable), it may be useful when some of the data is meaningless (i.e. a distracted user who is temporarily not paying attention), or when the parameters of the model may change over time.  Also, in practice, if $\lambda$ is large enough, the algorithm may be close ``enough" to an optimal solution.

{
Motivated by findings from online learning, we also expect that \textit{Follow-the-leader} (FTL) (or if regularization is used, \textit{Follow-the-regularized-leader} (FTRL, a.k.a. FoReL)) may improve regret rates relative to OGD, generally at the expense of additional computational cost  \cite{Shalev-Shwartz2011} (though without much computational burden if RLS can be used).  
We derive that} under mild conditions that hold for the SSKF learned with mild regularization, FTL obtains a regret rate of $\mathcal{O}(\log K)$ \cite{Hazan2007} (see Text S1 for details and discussion of constants).
Thus, keeping in mind these bounds are worst-case, we expect that using $FTL$ updates will provide improved performance relative to OGD or MA.  We validate {our theoretical results} in simulations in the next section.  We note that BCI datasets remain small enough that FTL updates for sets of reaches should be tractable, at least for initial decoder learning in closed-loop settings.

While the focus here is on static models, we note that there is additional literature concerning online optimization for \textit{dynamic} models. Here dynamic refers to situations where the neural tuning drifts in a random fashion over time.  Intuitively, something more like OGD is reasonable, and specific variants have been well characterized \cite{hall2015online}.
If the absolute total deviation of the time-varying parameters is constrained, these approaches can have regret of order $\mathcal{O}(\sqrt{K})$ \cite{hall2015online}.
A dynamic model may provide better fit and therefore provide lower MSE despite potential for additional regret.

\subsection*{Simulated cursor experiments}

\begin{figure}[!h]
\begin{center}
  \subfloat{\includegraphics[width=.35\textwidth]{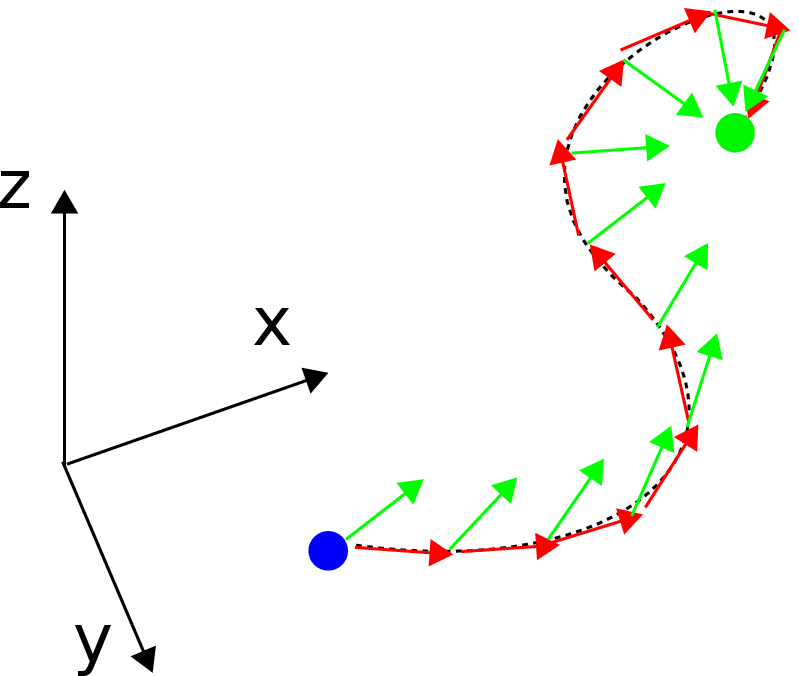}} 
  \hspace*{.5in}
  \subfloat{\includegraphics[width=.4\textwidth]{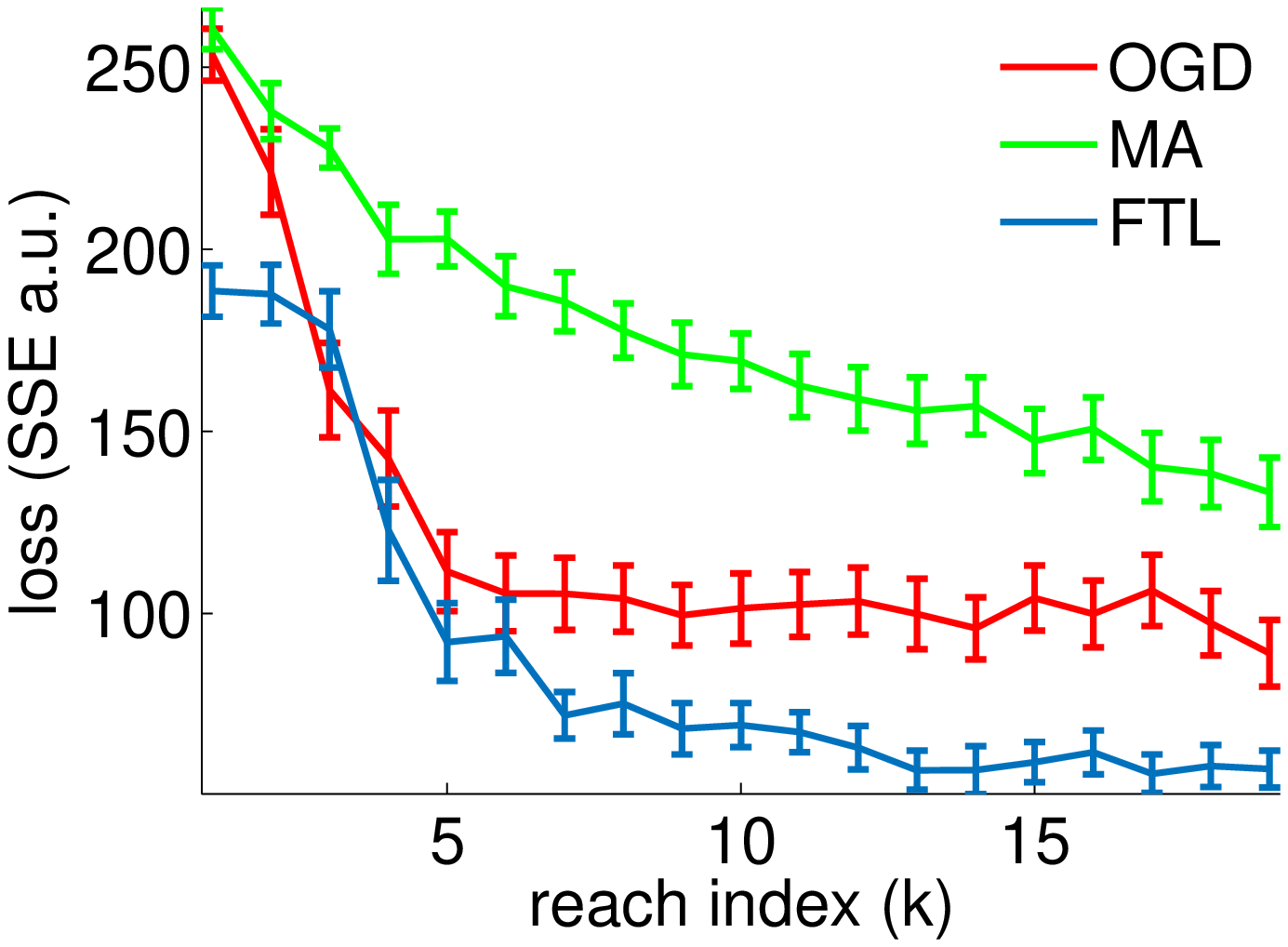}} 
  \end{center}    
  \captionsetup{width=.8\textwidth}
\captionsetup{font={footnotesize}}
\caption{Left panel is a cartoon of the cursor task.  The blue cursor is under user control and the user intends to move it towards the green target.  On a given reach trajectory, the cursor is decoded according to the current decoder yielding the path made up of red arrows.  At each state, the oracle intention is computed (green arrows) to be aggregated as part of $\mathcal{D}$ and incorporated into the update to the decoder.  In the right panel, we compare the performance of the algorithms on a simulation of the cursor task (loss incurred during each trial $k$).  We use Alg. \ref{alg_main} with the three update rules discussed (Alg. \ref{dec_updates} and Table \ref{table_regret}).  Intuitively, OGD makes less efficient use of the data and should be dominated by FTL.  Moreover OGD has additional parameters corresponding to learning rate which were tuned by hand. MA performs least well, though we selected $\lambda$ to be sufficiently close to 1 as to permit performance to gradually improve (smaller lambda leads to more unstable learning).  Each update index corresponds to the inclusion of 1 additional reach.  The entire learning procedure is simulated 100 times for each algorithm and errorbars are 2 standard errors across the simulations.}
\label{fig:cursor}
\end{figure}

The first set of simulations concerns decoding from a set of neurons that are responsive to intended movement velocity (see methods for full details).  In these simulations, there is a cursor that the user intends to move towards a target, and we wish to learn the parameters of the decoder to enable this. The cursor task (leftside panel of Fig. \ref{fig:cursor}) is relatively simple, but the range of results we obtain for well-tuned algorithm variants is consistent with our theoretically-motivated expectations.  Indeed, in the right panel of Fig. \ref{fig:cursor}, we see that the OGD algorithm, which takes only a single gradient step after each reach, performs less well than the FTL algorithm that performs batch-style learning using all data acquired to the current time.  MA performs least well, though for large values of $\lambda$ (i.e. .9 in this simulation), the performance can become reasonable.  

We also note that updates may require regularization to be stable, so we provide all algorithms with equal magnitude $\ell_2$ regularization (the regularization coefficient per OGD update was equal to $1/K$ times the regularization coefficient of the other algorithms).  After fewer than 10 reaches the OGD and FTL appear to plateau -- this task is sufficiently simple that good performance is quickly obtained when SNR is adequate.  { We note that we have opted to show sum squared error (SSE) rather than MSE (in Fig. \ref{fig:cursor} and elsewhere), because it reflects the aggregated single timestep error combined with differences in acquisition time -- MSE normalizes for the different lengths of reach trajectories, thereby only providing a sense of single timestep error (compare to Fig. S1).}

\begin{figure}[!t]
\begin{center}
  \subfloat{\includegraphics[width=.8\textwidth]{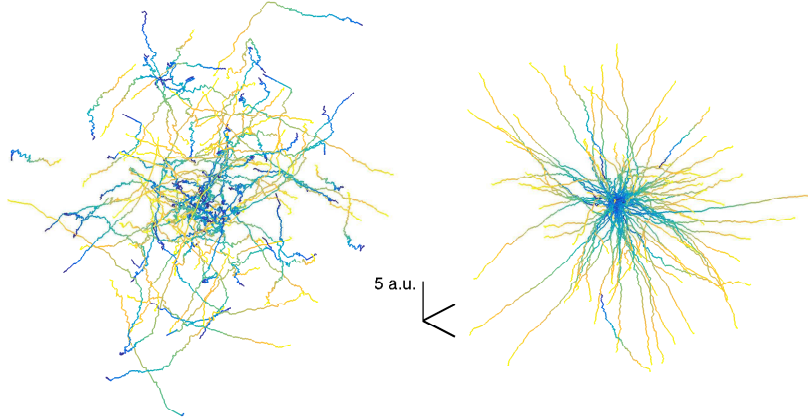}} 
  \end{center}    
  \captionsetup{width=.8\textwidth}
\captionsetup{font={footnotesize}}
\caption{Left panel is a visualization of 100 3D reach trajectories for a poorly-performing initial decoder (trained on 1 reach).  Right panel visualizes 100 trajectories for a well-performing decoder fit from 20 reaches (approximately at performance saturation for this level of noise).  Each trajectory is depicted with yellow corresponding to initial trial time and blue corresponding to end of trial (time normalized to take into account different reach durations).  The goals were in random locations, so to superimpose the set of traces, all positions have been shifted relative to the goal such that goal is always centered.  Observe that the initial decoder is essentially random and the learned decoder permits the performance of reaches which mostly proceed directly towards the goal (modulo variability inherited from the neural noise). Units here relate to those in Fig. \ref{fig:cursor} -- here referring to position as compared with MSE of corresponding velocity units.}
\label{fig:cursor_single}
\end{figure}

To get a sense of the magnitude of the performance improvements (i.e. the scale of the error in Fig. \ref{fig:cursor}), we can visualize poorly-performed reaches from early in training and compare these against well-performed reaches from a later decoder (Fig. \ref{fig:cursor_single}).  While the early decoder performs essentially randomly, the learned decoder performs quite well, with trajectories that move rapidly towards the target location.  See 
\href{https://drive.google.com/open?id=0B59a-51cjEx8d0lpSVg1WWhLZDg}{Mov. S1} 
for an example movie of cursor movements during the learning process.  

We emphasize that FTL essentially has no learning-related parameters (aside from the optional $\ell_2$-regularization coefficient).  On the other hand, OGD and MA have additional learning parameters that must be set, which may require tuning in practical settings.  The OGD experiments presented here are the result of having run the experiment for multiple learning rates and we reported only the results of a well-performing learning rate (since this requires tuning, it may be non-trivial to immediately achieve this rate of improvement in a practical setting where the learning rate is likely to be set more conservatively).  Too large a learning rate leads to divergence during learning, and too small a learning rate leads to needlessly slow improvement.

\subsection*{Simulated arm-reaching experiments}

\begin{figure}[!h]
\begin{center}
  \subfloat{\includegraphics[width=.4\textwidth]{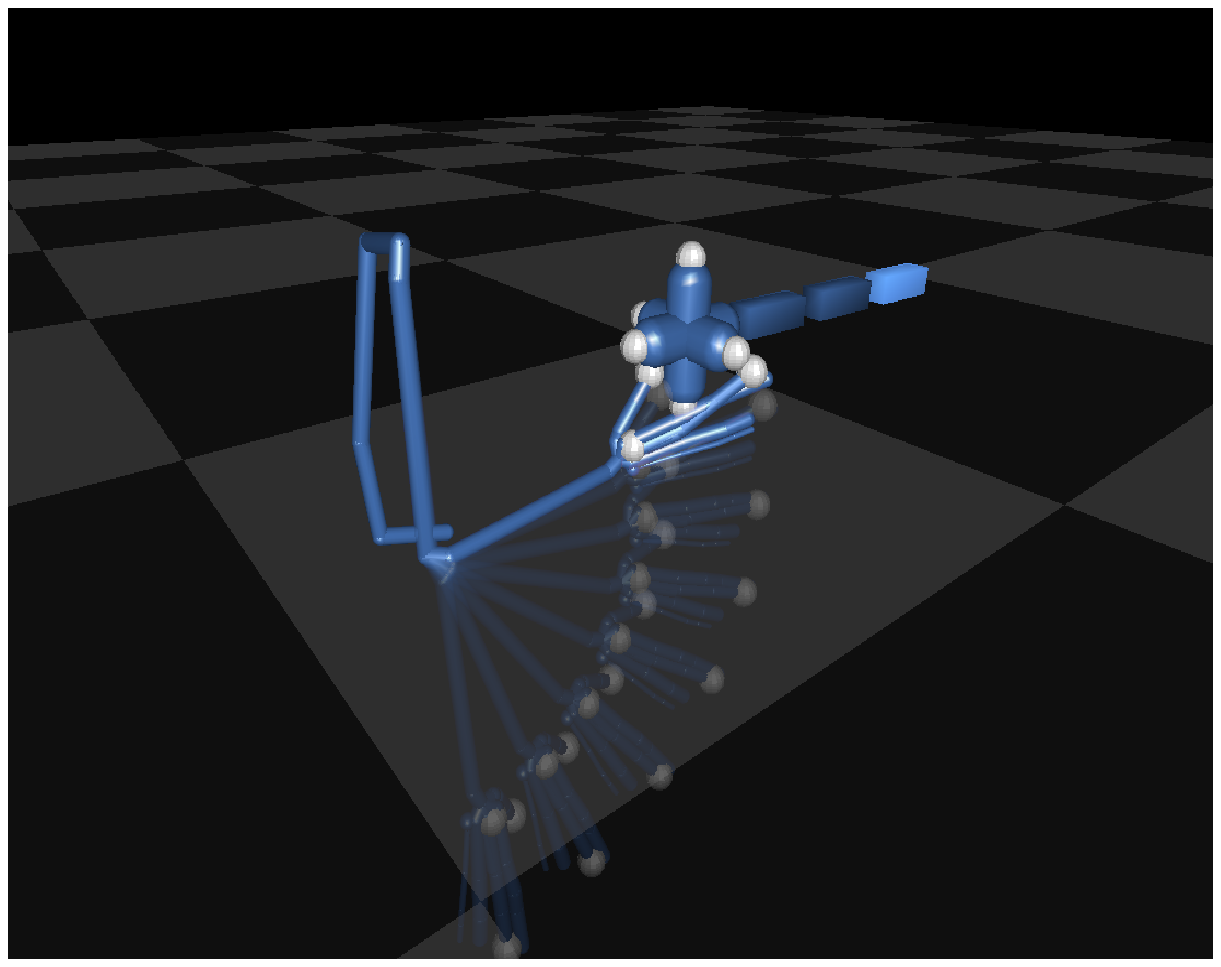}} 
  \hspace*{.5in}
  \subfloat{\includegraphics[width=.4\textwidth]{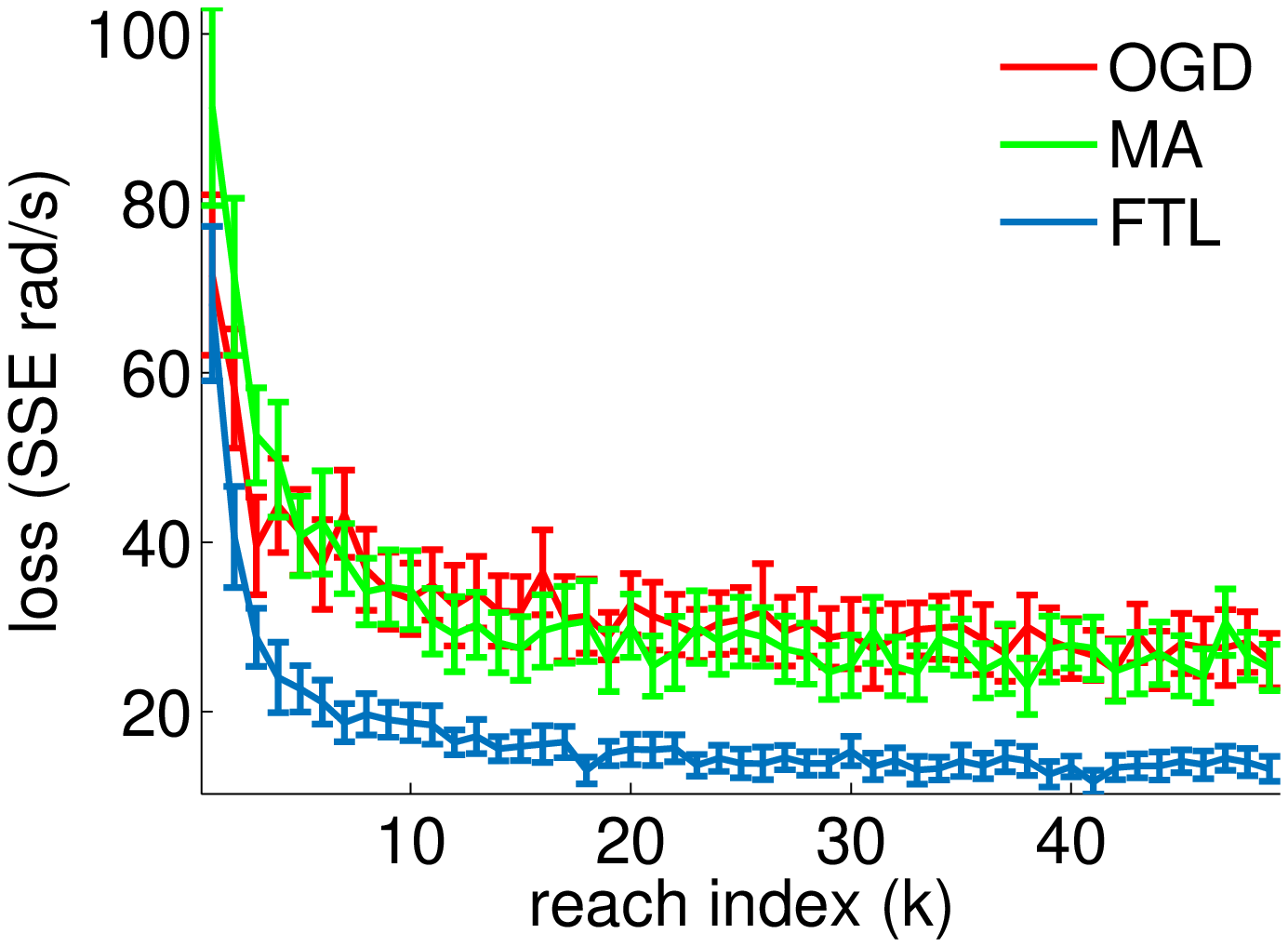}}   
  \end{center}    
  \captionsetup{width=.8\textwidth}
\captionsetup{font={footnotesize}}
\caption{Left panel depicts arm model in MuJoCo software and a trajectory of the arm during a simulated closed-loop experiment, after the decoder has learned to imitate the optimal policy (for illustration). This particular trajectory consists mostly of movement of an elbow joint, followed by slight movements of the middle finger and thumb when near the target. Right panel depicts a comparison of loss (here {SSE} of decoded joint angular velocities relative to oracle) as a function of reach index for the different update rules (similar to Right panel in Fig. \ref{fig:cursor}). In this plot, we consider only the loss for the shoulder, elbow, and wrist DOF as these are the dominant DOF (curves are similar when other critical joints are included). We see that FTL again gives good performance both in terms of rate of convergence and resulting solution (see Fig. \ref{fig:arm_single_trials} or 
\href{https://drive.google.com/open?id=0B59a-51cjEx8bUtPUHBaaDVibjg}{Mov. S2} 
for a sense of the quality of the performance). The entire learning procedure is simulated 50 times for each algorithm and errorbars are 2 standard errors across the simulations.}
\label{fig:arm}
\end{figure}

{In this section we introduce a new opportunity, moving beyond BCI settings where intention-based algorithmic capabilities have yet been explored. 
We validate }the imitation learning framework through simulation results on a high dimensional task -- BCI control of a simulated robotic/virtual-arm (Fig. \ref{fig:arm}).  Whereas existing algorithms {cannot be generalized to more complicated tasks, our results allow for generalization to an arm effector}. The simple ReFIT-style oracle of rotating instantaneous velocities towards the ``goal" is {ill-posed in general cases} -- the goal position could be non-unique and the different degrees of freedom (DOF) may interact nonlinearly in producing the end-effector position (both of these issues are present for an arm). Instead, we introduce an optimal control derived intention-oracle.  As our proof of concept, we present a set of simulated demonstrations of reaches of an arm towards a target-wand.  We envision this being incorporated into a BCI setting such as that described in \cite{Putrino201568}, where a user controls a virtual arm in a virtual environment.  Extension to a robotic arm is also conceptually straightforward, if a model of the robotic arm is available.  

For these simulations, we implement the reach task using a model of a rhesus macaque arm in MuJoCo, a software that provides a physics engine and optimal control solver \cite{todorov2012mujoco}. 
The monkey arm has 26 DOF, corresponding to all joint-angles at the shoulder, elbow, wrist, and fingers.  
The task objective we specified corresponds to the arm reaching towards a target ``wand," placed in a random location for each reach, and touching the wand with two fingers.  
Following from the task objective, at each timestep the optimal control solver receives the current position of the arm and the position of the goal (i.e. wand position), from which it computes incremental updates to the joint angles.  These incremental updates to the joint angles correspond to oracle angular velocities and we wish to learn a decoder that can reproduce these updates via Alg. \ref{alg_main}.   
See methods for complete details of the simulations.  

{Given that this arm task is ostensibly more complicated than cursor control, it may be initially surprising that we see that task performance rapidly improves with a small number of reaches (Right panel Fig. \ref{fig:arm}, and see 
\href{https://drive.google.com/open?id=0B59a-51cjEx8bUtPUHBaaDVibjg}{Mov. S2} 
for an example movie of arm reaches during the learning process).  
However, this relatively rapid improvement makes sense when we consider that the data is not collected independently, rather there is a closed-loop sequential process (see Alg. \ref{alg_main}).  
Consequently} we expect that early improvement should occur by leveraging the most widely used DOF (i.e. shoulder, elbow, and to a lesser extent wrist).  More gradually, the other degrees of freedom should improve (i.e. finger and less-relevant wrist DOF).  

\begin{figure}[!h]
\begin{center}
  \subfloat{\includegraphics[width=1\textwidth]{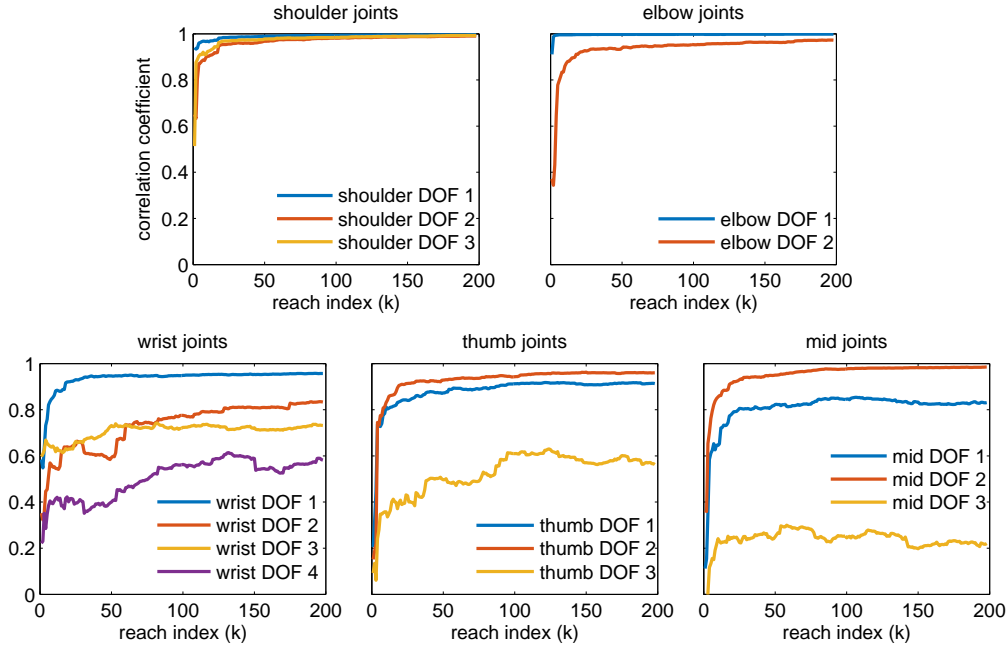}}  
  \end{center}    
  \captionsetup{width=.8\textwidth}
\captionsetup{font={footnotesize}}
\caption{Panels depict correlation between ``true" encoding model and estimated encoding model parameters as a function of index over reach trajectories (for a single trial). Each curve corresponds to the correlation for a different DOF.  The encoding model parameters are not directly guaranteed to converge.  We see, as expected, that the encoding model will improve for specific DOF in proportion to the extent to which those dimensions are relied on to perform the task.  Shoulder DOF are crucial for the task, being implicated in most reaches, so are learned rapidly.  Wrist and finger joints are relatively less critical for task performance, so are learned more gradually. In the thumb and middle finger panels above, the least well-learned DOF (thumb DOF 3 and mid DOF 3) can be interpreted as the ``distal inter-phalangeal joint" (i.e. the small joint near tip of the finger), which is not heavily relied upon in this reach task.}
\label{fig:arm_encoding}
\end{figure}

To empirically examine the rate at which we can learn about distinct DOF, we conduct an analysis to see how well we can characterize the mapping between intention (per DOF) and neural activity.  
At each stage of the learning process ($k=1...K$), we use the aggregated dataset $\mathcal{D}^{(1:k)}$ to estimate the encoding model by regression (see methods, Eqn. \ref{obs_eqn_main}).  The encoding model corresponds to the mapping from intention to neural activity and our ability to recover this (per DOF) reflects the amount of data we have about the various DOF.   
To quantify this, we compute correlation coeffcients (per DOF, across neurons) between the true encoding model parameters (known in simulation) and the encoding model parameters estimated from data aggregated up through a given reach.  
We expect this correlation to generally improve with increasing dataset size; however, regret bounds do not provide direct guarantees on this parameter convergence.  
The key empirical observation is that DOF more integral to task performance are learned rapidly, whereas certain finger DOF which are less critical are learned more gradually (Fig. \ref{fig:arm_encoding}). 

\begin{figure}[!h]
\begin{center}
  \subfloat{\includegraphics[width=.4\textwidth]{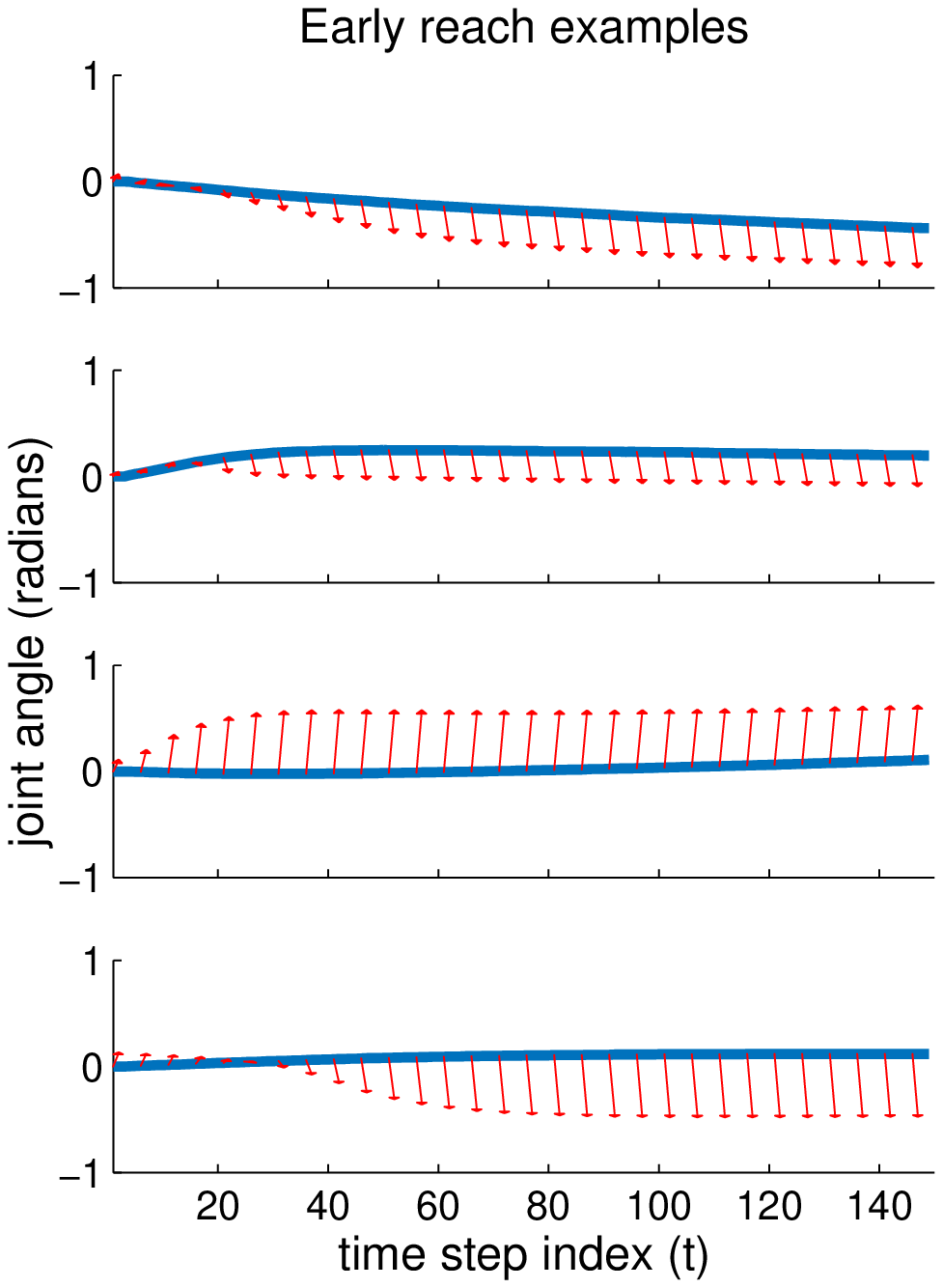}} 
  \hspace*{.35in}
  \subfloat{\includegraphics[width=.4\textwidth]{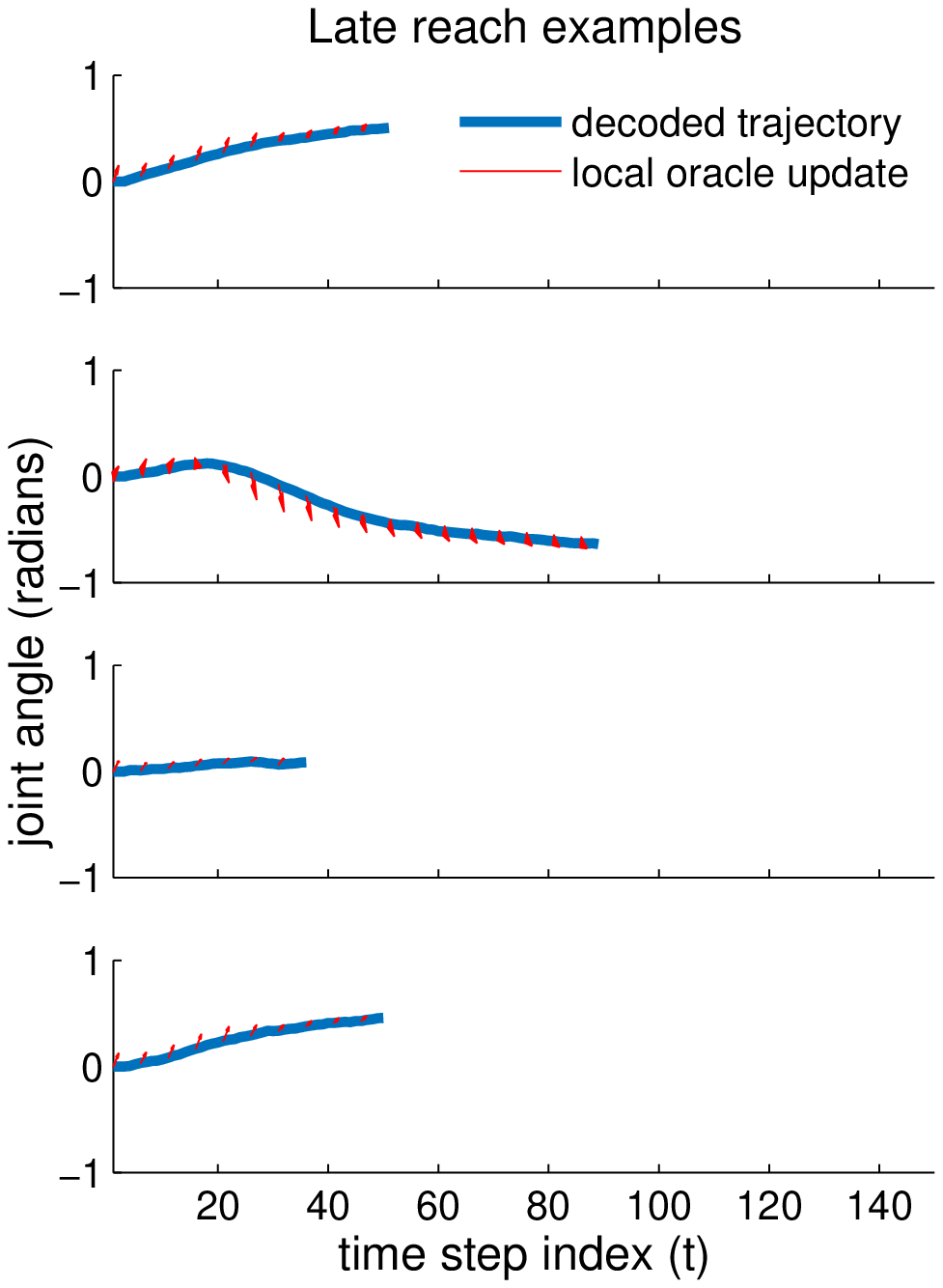}} 
  \end{center}    
  \captionsetup{width=.8\textwidth}
\captionsetup{font={footnotesize}}
\caption{
Plots depict reach trajectories of a representative shoulder DOF for 4 paired examples of reaches, from separate re-initializations of the decoder (i.e. different trials).  
Left panels show a poorly-performing early decoder (k=2), and right panels show a well-performing decoder (k=30).  Rows correspond to matched pairs of reaches for different repeats of the experiment.     
Blue curves correspond to the actual decoded pose of the DOF over time, and red arrows depict the local oracle update (only visualized for a subsampling of timesteps). 
For the early reaches, observe that the decoder does not always proceed in the intended direction.  
For the late reaches, observe that actual pose updates are quite consistent with the oracle and trajectories are shorter because the targets are acquired more frequently and more rapidly.}
\label{fig:arm_single_trials}
\end{figure}

Similarly to the cursor tasks, we want to examine the magnitude of the performance improvements.  For this case, it is difficult to statically visualize whole reaches. Instead, we look at an example shoulder DOF and depict the trajectory of that joint during a reach (Fig. \ref{fig:arm_single_trials}).  Branching off of the actual trajectory, we show local, short-term oracle trajectories which depict the intended movement.  Note that the oracle update takes into account other DOF and optimizes the end-effector cost, so it may change over time as other DOF evolve.  We see that the early decoder does not yield trajectories consistent with the intention -- the decoded pose does not move rapidly, nor does it always move in the direction indicated by the oracle.  The late decoder is more responsive, moving more rapidly in a direction consistent with the the oracle.  In the four examples using the late decoder, the arm successfully reaches the target, so the reach concludes before the maximum reach time.  

{

\subsection*{Model mismatch}

An important potential class of model mismatch arises when there is a discrepancy between the ``oracle" policy and the true intention of the user (in such cases the oracle is not a proper oracle and is better thought of as an attempt at approximating an oracle).  We can consider this setting to suffer from ``intention mismatch" (see \cite{golub2015internal} for a distinct, but related concept of discrepant ``internal models"). 

In our basic simulations, we have assumed we have a true intention oracle.  When such an oracle is available, this represents the ideal statistical setting and our simulations provide a sense of quality of algorithmic variants in this setting.  However, we next consider the realism of this assumption and the consequences if it is not correct.  There are a few classes of deviations we might expect between a true user's intention and the intention oracle.  

One class of intention mismatch occurs when the actual user intention and the oracle are related by a fixed linear operator (e.g. fixed rotation).  In this case, the learned decoder will produce the same kinematics as the case without mismatch (modulo SNR concerns). Although the loss between actual user intention and decoded kinematics will be increased, the resulting decoder follows the oracle and matches the BCI performance of the zero mismatch case (verified in simulations, not shown). 

\begin{figure}[!t]
\begin{center}
  \subfloat{\includegraphics[width=.4\textwidth]{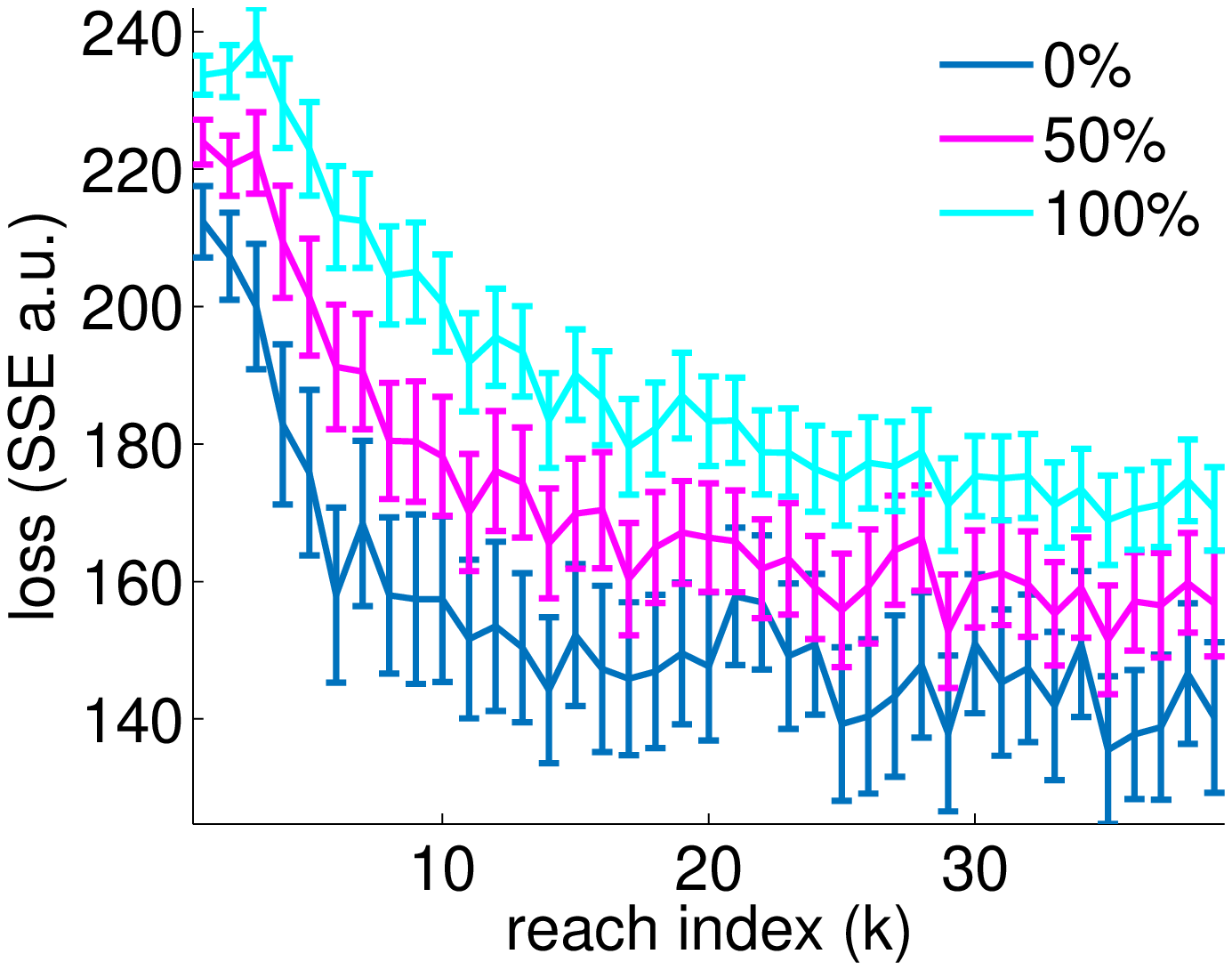}} 
  \hspace*{.35in}
  \subfloat{\includegraphics[width=.4\textwidth]{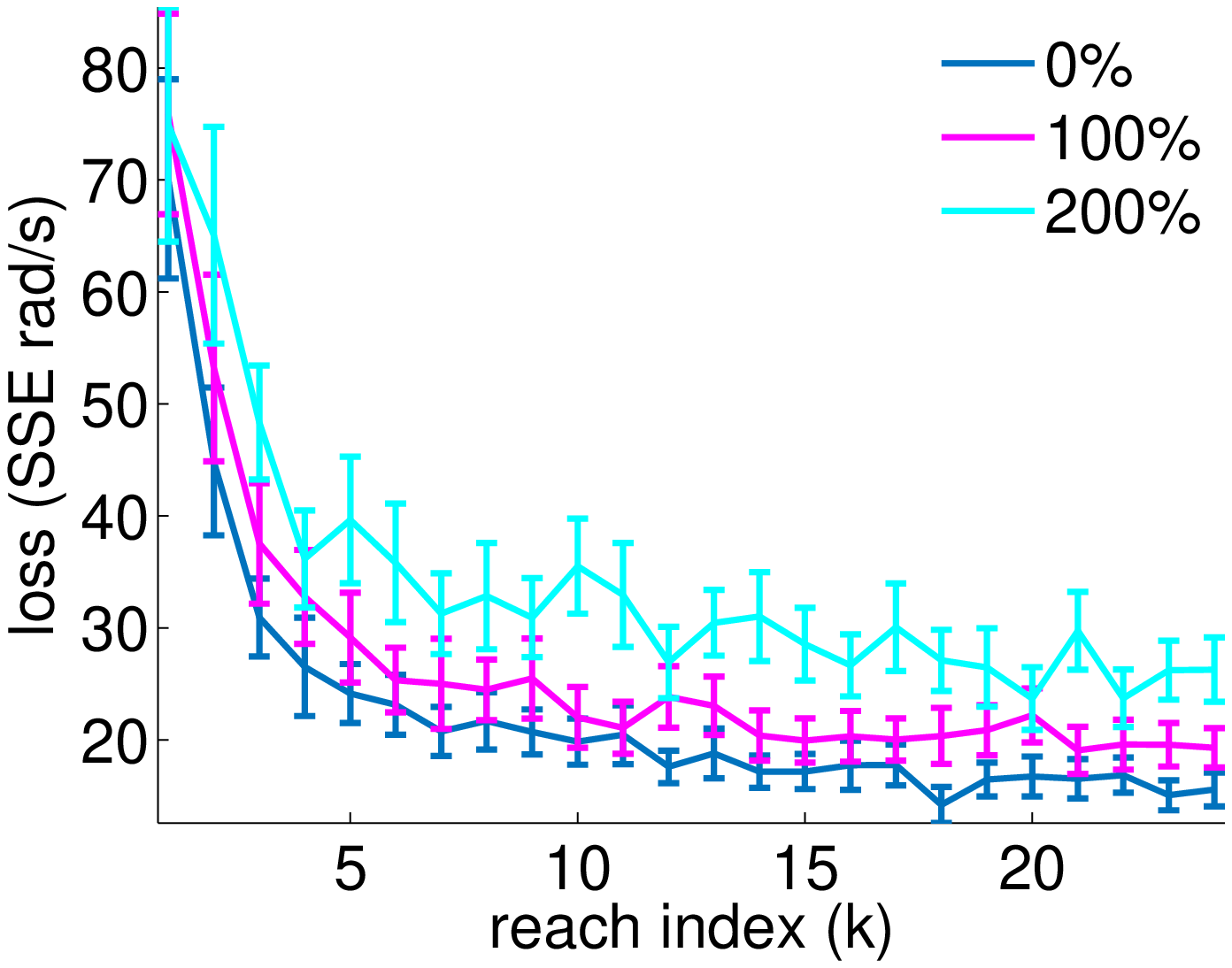}} 
  \end{center}    
  \captionsetup{width=.8\textwidth}
\captionsetup{font={footnotesize}}
\caption{
Plots depict decline in performance (i.e. loss between noise-free oracle and decoded intention) with intention noise model mismatch using sum square error (SSE) over the duration of a reach for (left) cursor task and (right) arm reaching task trajectories, comparable to performance curves in Figs \ref{fig:cursor} and \ref{fig:arm} respectively.  In each task, noise performance curves are obtained when the user's intent is a noisy version of the oracle, captured by a linear combination of intention oracle and a random vector.  The noise level is indicated by a noise percentage, corresponding to the magnitude of the noise relative to the intention oracle signal.}
\label{fig:model_mismatch_noise}
\end{figure}

A second class of intention mismatch corresponds to random noise applied to the user intention that is not known by the oracle.  For example, it is straightforward to assume the the user's actual intention in a single timebin is inherently variable due to noise inherited from sensory feedback noise, or due to inherent variability in biological control.  For such a case, we perform simulations identical to those performed previously, but we model the actual user intention (that drives the simulated neural activity) as a combination of a random intention and the oracle intention.  The magnitude of the intention noise here corresponds to the magnitude of the random intention relative to the oracle (i.e. 100\% noise indicates that actual user intention is a linear combination of the oracle intention and a random vector of equal magnitude).  We emphasize that here the oracle is not correct and there is additional noise in the system that is from the random intention.  We can verify empirically that there is not catastrophic failure with this intention noise variant of model mismatch (see Fig \ref{fig:model_mismatch_noise}).
Naturally, performance (i.e. loss between noise-free oracle and decoded intention) decreases when there is additional noise.  A prominent feature is that the decline in performance is graceful rather than catastrophic.

The final class of intention mismatch occurs when the oracle is systematically and nonlinearly distinct from the oracle.  Intuitively this would mean that the oracle is systematically and reliably wrong in a state dependent way -- i.e. the direction the oracle is wrong, depends upon the current pose.    

}

\section*{Discussion}
 
{In this work, we have unified closed-loop decoder training approaches by providing a meta-algorithm for BCI training, rooted in imitation learning.}
Specifically, we have focused on the parameter learning problem, complementing other research that focuses on the problem of selecting a good decoder family \cite{zhang2015recasting}.  
Our approach allows the parameter learning problem to be established on a firmer footing within online learning, for which theoretical guarantees can be made.  
This is crucial since ReFIT-based approaches are being translated to human clinical applications where performance is of paramount concern \cite{gilja2015clinical,jarosiewicz2015virtual}. 
Moreover, we have demonstrated that this approach now permits straightforward extension to higher dimensional settings, enabling rapid learning even in the higher dimensional case. 
In scaling existing algorithms to an arm-control task, we have provided generic approaches to solve two issues.  
First, imitation learning (using data aggregation) serves as the generic framework for updating parameters.  
Second, we have employed a generic, optimal control approach, which can be used to compute intention-oracle kinematics in a broad range of BCI settings.  

{
For simulations in this work, we employ linear encoding of kinematic variables because, in addition to having a history in the BCI literature \cite{wu2006bayesian}, this corresponds to an operationally useful encoding model employed in recent, well-performing applications in the closed-loop BCI \cite{gilja2012high,gilja2015clinical}.  
We do not intend to claim that simple, linear encoding models as assumed when employing Kalman filter decoders correspond to the reality of innate neural computation in motor cortex.  
Nonlinear filtering approaches that make more realistic assumptions about neural encoding have been explored offline \cite{shoham2005statistical,wang2009sequential,nazarpour2012emg}.  However, it is not clear that offline results employing more realistic encoding models always translate performance gains to closed-loop settings \cite{koyama2010comparison}.  Nevertheless, there have been successes using more complicated decoding algorithms in closed-loop experiments \cite{hagai,li2009unscented,sussillo2012recurrent}.
Following on recent scientific work that has sought to understand a role of intrinsic dynamics in motor cortices \cite{churchland2012neural}, dynamics-aware decoders are also being developed \cite{wu2009neural,lawhern2010population,kao2015single}.
While many decoder forms may be considered, in line with the variety of theories about the motor cortex, the precise choice is orthogonal to the work here.
Intention-based parameter fitting does not depend, in any general way, on the encoding model assumed by the decoding algorithm.    
Consequently, a key benefit of the theoretical statements we present are that the algorithm performance guarantees hold for general classes of decoders, and the meta-algorithm we describe is largely agnostic to the details of the encoding.  
}

It is a key point that Alg. \ref{alg_main} results in preferential acquisition of data that enables learning of the most task-relevant DOF.  
This follows from the fact that the sampling of states in closed-loop is non-uniform, since the current decoder induces the distribution of states visited during the next reach.
Exploration is not explicitly optimized, but more time is spent in relevant sets of states as a consequence of preferential sampling of certain parts of what can be a high dimensional movement space.
This clarifies the potential utility of assisted decoding, which may serve to facilitate initial data collection in positions in the movement space that are especially task-relevant. 
{
This non-uniform exploration of the movement space provides intuition for the generality of the theoretical guarantees for \textsc{DAgger}-like learning. The decoder used in this work is of a relatively simple form (\textit{steady-state velocity Kalman filter}, described in methods), but the theoretical results hold for general stationary, deterministic decoders.
}

While we have focused on a simple, parametric decoder, the parameter learning approach presented in this paper extends to more complicated decoders.
For example, we may wish to allow the neural activity to be decoded differently depending on the current state of the effector.  
{
In conventional imitation learning, policies are trained to yield sequences of actions (without user input), so this general problem is extremely state-dependent.
By building into the decoder an expressive mapping that captures state-transition probabilities, we could design a policy-decoder hybrid to exploit regularities in the dynamics of intended movements and heavily regularize trajectories based on their plausibility. 
}
Similarly, this framework accommodates decoders which operate in more abstract spaces (such as if the available neural activity sent action-intention commands rather than low-level velocity signals).  

A particularly interesting opportunity { that corresponds to an augmentation of \textit{follow-the-leader} (FTL, eqn. \ref{FTL_eq}) would be to enrich} the decoder family as the dataset grows.  We can imagine a system with decoders of increasing complexity (more parameters or decreasing regularization) as the aggregated dataset of increasing size becomes available.  While we focused on a simple decoder (i.e. the Kalman filter) which makes sense for small-to-moderate datasets, some work suggests that complicated decoders trained on huge datasets can perform well (e.g. using neural networks \cite{sussillo2012recurrent}).  
We anticipate that data aggregation would allow us to start with a simple decoder, and we could increase the expressive power of the decoder parameterization as more data streams in. 
   
{Our formalization of BCI learning most closely resembles the \textsc{DAgger} setting, but novel extensions to the BCI learning setting follow from related imitation learning formulations.
}
Some particularly relevant opportunities are surveyed here.  When starting from an initial condition of an unknown decoder-policy, it may be hard to directly train towards an optimal decoder-policy.  Training incrementally towards the optimal policy via intermediate policies has been proposed\cite{he2012imitation}.  Under such a strategy, a ``coach" replaces the oracle, and the coach provides demonstration actions which are not much worse than the oracle but are easier to achieve.  For example, in BCI, it may be hard to learn to control all DOF simultaneously, so a coach could provide intention-trajectories that use fewer DOF.  It has also been observed that \textsc{DAgger} explores using partially optimized policies, and these might cause harm to the agent/system.  Especially early in training, the policies may produce trajectories which take the agent through states which may be dangerous to the agent or the environment.  An appropriate modification to solve this is to execute the oracle/expert action at timesteps when a second-system suspects there may be an issue carrying out the policy action, thereby promoting safer exploration \cite{kim2013maximum}.  

{
As touched upon in the results, we also want to be aware of the performance impact of model mismatch and mitigate this problem.
We speculate that generic one-step intention mismatch may be well modelled by random deviations, because when the problem consists of sequential decisions, it is not clear how systematic discrepancies between the one-step-ahead oracle and the one-step-ahead intention would arise.  As such, so as long as the intention mismatch is moderate, its effect on learning may be no worse than the effect of the noise variant of intention mismatch characterized in the results.  
Regardless, in such settings where, even after carefully designing the intention oracle there is persistent mismatch, a combined imitation learning and reinforcement learning approach may produce better results \cite{kim2013learning}.  
This amounts to a hybrid optimization that combines the error-ridden expert signals with RL signals obtained by successful goal acquisitions.}
 
Finally, in this work we have assumed there is not gradual ``drift" in the neural encoding model -- it is probably a fair assumption that neural encoding drift is not a dominant issue during rapid training \cite{chestek2011long,kowalski2013dynamic}.  We highlight a distinction between general closed-loop adaptation (where the decoder should adapt as fast as possible), versus settings designed for the user to productively learn, termed co-adaptive (for a review of co-adaptation, see \cite{shenoy2014combining}).  We have focused on the setting with user learning in other work \cite{multiagent,merel2015encoder}, but we here focused on optimizing parameter learning under the assumption that the user's neural tuning is fixed, allowing us to rigorously compare algorithms.  In future work, it may prove fruitful to attempt to unify this analysis with co-adaptation.  We also anticipate future developments that couple the sort formalization of decoder learning explored in this work with more expressive decoders.  We are optimistic that progress in these directions will enable robust, high-dimensional brain-computer interface technology.

\section*{Methods}

\subsection*{Simulated experiments}

In this work we present two sets of simulations.  The first set of simulations consist of simulated closed-loop experiments of 3D cursor control.  In these simulations, the cursor serves as the \textit{effector}, and this cursor is maneuverable in all three dimensions.  Goals are placed at random locations and the \textit{task objective} is to minimize the squared error loss (mean squared error, MSE; summed squared error, SSE) between the cursor and the current goal.  Goals are acquired when the cursor is moved to within a small radius of the target.  The \textit{oracle} for this task is determined from optimal control.  When there is a quadratic penalty on instantaneous movement velocity, the optimal trajectory from the cursor towards the goal will be equal-length vectors directed towards the target.  So at each timestep, we take the oracle to correspond to a goal-directed vector from the current cursor position.     

The second set of simulations are similar, but involve controlling an arm to reach towards a ``wand".  As the \textit{effector}, we use an arm model with dimensions corresponding to those of a rhesus macaque monkey used for BCI research 
(from Pesaran Lab, Center for Neural Science, New York University, http://www.pesaranlab.org, as in \cite{Putrino201568}). For simplicity we treat each joint as a degree of freedom (DOF) yielding 26 joint angles and 26 corresponding angular velocities.  We specify the \textit{task objective} to be a spring-like penalty between the wrist position (3D spatial coordinates) and the wand position.  Specifically, in addition to the 26 joint-angle DOF, there are also identifiers corresponding to the x-y-z coordinates of the wrist and select fingertips, as well as points on the wand.  Objective functions in terms of the x-y-z coordinates of these markers can be specified, and the MuJoCo solver computes trajectories (in terms of the specified joint angles) in order to optimize the objective.  We defined the initial objective in terms of the Euclidean distance between the wrist and the wand.  Once the wrist is within a radius $\delta$ of the wand, a new sping-like penalty is placed on the distance between tip of the middle finger and a point on the wand and also between the tip of the thumb and a point on the wand -- this causes the fingertips to touch two points of the wand (a simple ``grasp").  This explicit task objective allows us to compute the \textit{oracle} solution for the reach trajectory, and this oracle is computed via an iterative optimal control solver on all joint angles in the model.  The model, simulation, and optimal control solver are implemented in an early release of the software simulation package MuJoCo \cite{todorov2012mujoco}.  
At each timestep, given the wand position and current arm position, the optimal control solver produces an incremental update to all (26) of the joint angles of the arm, and this goal-directed angular velocity vector is taken as the oracle. 
As an alternative to explicitly posing the objective function and computing the oracle, one can imagine using increasingly naturalistic optimal-control-based oracles that use more elaborate motor models trained on real data \cite{Berniker2015}.

We produce synthetic neural activity similarly in both sets of simulations.  In the cursor task, we want the neurons to be tuned to intended cursor velocity.  In the arm task, neurons should encode velocities of the joint angles.  
To produce simulated neural data that reflects the ``user's" intention, we have a convenient choice -- the oracle itself.  
The simulation cycle entails: (1) computing the intention-oracle (given the current state, goal, and task objective), (2) simulating the linear-Gaussian neural activity from the intention-oracle (Eqn. \ref{obs_eqn_main}), (3) using the current decoder to update the effector, and (4) updating the decoder between reaches.
We note that the oracle is used twice, first to produce the neural activity and subsequently in the imitation learning decoder updates. 

Specifically, we simulate neural activity via the neural encoding matrix $A$ that maps intended velocity to neural activity:
\begin{equation}
\nv_t = A \ov_t + \textbf{c}_t \text{ (with $\textbf{c}_t \sim \mathcal{N}(0,\textbf{C})$)}. \label{obs_eqn_main}
\end{equation}
where the noise covariance $C$ was taken to be a scaled identity matrix, such that the signal-to-noise ratio (SNR) was $\approx1$ per neuron (i.e. noise magnitude set to be roughly equal in magnitude to signal magnitude per neuron, which we considered reasonable for single unit recordings). 
In real settings this neural activity might be driven by some intended movement $\xv_t^*$ (where here the star denotes intention as in \cite{zhang2015recasting}).

These simulations assume the intention-oracle is ``correct".  As such, a feature of all ReFIT-inspired algorithms is that there is model mismatch if the user is not engaged in the task or has a meaningfully different intention than these algorithms presume.  This problem affects any algorithm that trains in closed-loop and makes assumptions about the user's intention (see discussion for extensions to handle the case when the oracle is known to be imperfect).  

For the simulations, $A$ was selected to consist of independently drawn random values.  For both tasks, we randomly sampled a new matrix $A$ for each repeat of the simulated learning process.  For the cursor simulations, we simply sampled entries of $A$ independently from a normal distribution.  For the higher dimensional arm simulations, we wanted to have neurons which did not encode all DOFs, so for the results presented here we similarly sampled $A$ from a normal distribution, but then set any negative entries of $A$ to zero (results were essentially the same if negative entries were included).   

Assisted decoding (see Alg. \ref{alg_main}) was not heavily used.  To provide stable initialization, $\beta_0$ was set to 1 (and noise was injected into the oracle for numerical stability), and all subsequent $\beta_k$ were set to 0. For the cursor simulations, we used 10 neurons and the maximum reach time $T$ was set to 200 timesteps.  For arm simulations, we used 75 neurons and the maximum reach time $T$ was set to 150 timesteps.  We consider simulated timesteps to correspond to real timesteps of order 10-50ms.
  
For both sets of simulated experiments the \textit{decoding algorithm} was chosen to be the \textit{steady-state velocity Kalman Filter} (SSVKF), which is a simple decoder and representative of decoders used in similar settings (i.e. it corresponds to a 2nd order physical system according to the interpretation in \cite{zhang2015recasting}).  The SSVKF has a fixed parametrization as a decoder, but it also has a Bayesian interpretation.  When the encoding model of the neural activity is linear-Gaussian with respect to intended velocity, the velocity Kalman filter is Bayes-optimal, and the steady state form is a close approximation for BCI applications.

The \textit{steady state Kalman Filter} (SSKF) generally has the form:
\begin{equation}
\xv_{t+1} = {\bf F} \nv_t + {\bf G} \xv_t
\end{equation}
Here ${\bf G}$ can be interpreted as a prior dynamics model and ${\bf F}$ can be interpreted as the function mediating the update to the state from the current neural data.  In practice, a bias term can be included in the neural activity to compensate for non-zero offset in the neural signals.  The generic SSKF equation can be expanded into a specific SSVKF equation, where the state consists of both position and velocity.  At the same time we will constrain the position to be physically governed by the velocity, and we will only permit neural activity to relate to velocity.

\begin{equation}
\begin{bmatrix}
    {\bf p}_{t+1}\\
    {\bf v}_{t+1}
\end{bmatrix}
=
\begin{bmatrix}
    \textbf{0}      & \textbf{0} \\
    \textbf{F}_v & \textbf{b}_v
\end{bmatrix}
\begin{bmatrix}
    \nv_t\\
    \textbf{1}
\end{bmatrix}
+
\begin{bmatrix}
    \textbf{I}      & dt\times \textbf{I} \\
    \textbf{0} & \textbf{G}_v
\end{bmatrix}
\begin{bmatrix}
    {\bf p}_t\\
    {\bf v}_t
\end{bmatrix}
\end{equation}
    
It is straightforward to augment the decoder to include past lags of neural activity or state.  A very straightforward training scheme that is apparent for this specific decoder is to simply perform regression to fit $\{ \textbf{F}_v, \textbf{b}_v, \textbf{G}_v \}$, from the function:
\begin{equation}
\label{decode_equation}
\textbf{v}_{t+1} = \textbf{F}_v \textbf{n}_t + \textbf{b}_v + \textbf{G}_v \textbf{v}_t + {\bf e}_t
\end{equation}   
where ${\bf e}_t$ denotes an additive Gaussian noise term.

\section*{Supporting Information}

\subsection*{S1 Text}
\label{S1_Text}
{\bf Restatement of theoretical results for \textsc{DAgger} and presentation of regret bounds in linear-quadratic case.} We restate and interpret the theoretical results for \textsc{DAgger}.  We also describe specific bounds on regret for selected BCI update rules under a quadratic loss and linear decoder.

\subsection*{\textsc{DAgger} guarantees}

A major contribution of the \textsc{DAgger} meta-algorithm for the imitation learning setting is that the loss can be bounded in relation to online learning regret bounds \cite{Ross2011}.  This allows the analysis of \textsc{DAgger} through established results in the literature. Note that $d_\pi$ is the distribution over states that arise from policy/decoder $\pi$.

\begin{theorem}[modified statement of Theorem 4.1 from \cite{Ross2011}\label{thm:dagger}]
For \textsc{DAgger}, there exists a policy $\hat{\pi}\in \pi^{(1)},\dots{\pi}^{(K)}$ s.t. $\emph{$\mathbb{E}_{\xv,\nv,\ov \sim d_{\hat{\pi}}}$} [\ell(\hat{\pi}$\emph{($\xv,\nv),\ov$}$)]\leq \epsilon_K+\frac{1}{K}\cdot (2 \ell_{\max}) \left[T\sum_{k=1}^K\beta_k\right]+\gamma_K$, where $\gamma_K$ is the average regret of $\pi^{(1)},\dots{\pi}^{(K)}$.
\end{theorem} 
This theorem states that the expected loss of the policy (i.e. decoder) resulting from \textsc{DAgger} is upper bounded by three terms.  
The first and second terms depend on the model and meta-algorithm settings, whereas $\gamma_K$ will depend on the specific update method chosen.
We note as well that this assumes that T is an upper bound on $T_1,\dots,T_K$, so that the length of any trajectory is less than the maximum duration $T$.

We first discuss the two terms shared across update rules.  The first term:
\begin{equation}
\epsilon_K=\min_{\pi\in\Pi}\sum_{k=1}^K\mathbb{E}_{\xv,\nv,\ov \sim d_{\pi^{(k)}}}[\ell(\pi(\xv,\nv),\ov)],
\end{equation}
is the loss that would have been incurred if the best decoder had been used the whole time.  In some settings $\epsilon_K$ may approach 0, but typically there is observation noise (i.e. neural noise in this case), and additional variability could arise from model mismatch.  For BCI, we consider $\epsilon_K$ to be the error due to neural noise of the best decoder.
The second term is dependent on $\beta_i$, which controls how the algorithm blends the oracle policy and the learned decoder during training. $\beta_i$ can be chosen to be 0 for all $i$, which would eliminate this term.
However, this term decays quickly, $\mathcal{O}(\frac{1}{K})$, so assisted training may help performance in practice.
The constant $\ell_{\max}$ is the maximum value of $\ell(\pi(\xv,\nv),\ov)$ for $\pi\in \pi^{(1)},\dots,\pi^{(K)}$ and $k \in 1 ,\dots, K$ over effector states and neural observations.  
In the BCI setting, degrees of freedom of the effector have a bounded range and neural activity is physiologically bounded.  So we expect $\ell_{\max}$ will scale with the variance of the decoded variable.

The $\gamma_K$ term, which is the average regret of the $\pi^{(1)},\dots{\pi}^{(K)}$, is given by
\beq
\gamma_K=\text{Regret}_K(\Pi)/K.
\eeq

{
We note that Theorem \ref{thm:dagger} covers the asymptotic case.  Theorem 4.2 of \cite{Ross2011} addresses the finite sample case, where the bound will hold with probability at least $\delta$ when the term $\ell_{\max} \sqrt{\frac{2\log(1/\delta)}{K}}$ is added to the right hand side.  The effect of this is that it adds a small amount of slackness to Theorem 1 that increases if we want the theorem to hold with higher probability.  This term decays $\mathcal{O}\left(\frac{1}{\sqrt{K}}\right)$.  While this term may have a slower rate than the other terms, we note both that this will still maintain a no-regret algorithm and that for practical values this term will be comparatively small.
}

{
As written in \cite{Ross2011}, Theorem \ref{thm:dagger} requires the use of probabilistic mixing of the oracle policy and the learned policy.  Probabilistic mixing refers to using the oracle policy exactly with probability $\beta_k$ and the decoded policy exactly with probability $1-\beta_k$. However, we have advocated using a linear mixture of the decoding policy rather than probabilistic mixing to match the existing BCI literature.  We note that this will only superficially alter the form of Theorem \ref{thm:dagger}.  The term $\frac{1}{K}\cdot (2 \ell_{\max}) \left[T\sum_{k=1}^K\beta_k\right]$ is dependent on Lemma 4.1 of \cite{Ross2011}, but a slightly different bound can be derived using triangle inequality (the naive bound given in Lemma 4.1 of \cite{Ross2011}).  When using this linear mixing, the bound in Theorem \ref{thm:dagger} would instead have a term $\frac{1}{K}\cdot (2 \ell_{\max}) \left[\sum_{k=1}^K1_{\{\beta_k\neq 0\}}\right]$, where $1_{\{\cdot\}}$ denotes an indicator function (1 if the condition is true, 0 otherwise).  This term is the same for any non-zero $\beta_k$ chosen.  This is a very loose bound, but the original bound is also very loose on this term and for the sequences of $\beta_k$ used in the experiments there will be minimal differences.  For example, the sequence of $\beta_1=1$ and $\beta_k=0,~k>1$ will yield the same result on this term for linear mixing and probabilistic mixing.
}

\subsection*{Regret in linear-quadratic setting}
For a given objective function, we can use established results to analyze the decoder update options.
We specialize our statements for the case of a quadratic loss.  
The SSKF takes a linear autoregressive form, so concrete statements about the quadratic loss will be applicable for the SSKF decoder.  We consider a linear decoder which attempts to estimate intention from covariates -- we let ${\bf W}$ be the parameters of the linear decoder $\pi$ and $\textbf{z}_{kt}$ be the covariates.  With a slight abuse of notation, we define:
\begin{equation}
 \ell({\bf W},\textbf{z}_{kt},\ov_{kt})=||{\bf W} \textbf{z}_{kt}-{\bf o}_{kt}||^2.
\end{equation}
 This generic linear decoder may be explicitly specialized to the SSVKF in 
 Eqn. 8 
 by setting ${\bf W} = [\textbf{F}_v ~ \textbf{b}_v ~ \textbf{G}_v]$ and $\textbf{z}_{kt} = [\textbf{n}_{kt} ~ \textbf{1} ~ \textbf{x}_{kt}]^\top$.
 
The linear-quadratic setting is widely seen in applications { and mean square error convergence properties of linear models have been analyzed specifically for the least mean square (LMS) algorithm \cite{Widrow1985}}.  Perhaps surprisingly, the quadratic loss does not satisfy the assumptions required by the simplest online optimization frameworks { for regret analysis}, because the total loss is not a Lipschitz function \cite{Shalev-Shwartz2011}.  An $L$-Lipschitz function is defined as $|f(\xv+{\bf \delta})-f(\xv)|\leq L ||{\bf \delta}||$ with respect to a given norm, typically the $\ell_2$ norm (i.e. $||\xv||_2=\left(\sum_i x_i^2\right)^{1/2}$).  For a squared loss, $L$ would go to infinity at the tails.   However, the squared loss is a Lipschitz function over a bounded region, so in practical settings this is sufficient.

We consider the three updates in 
Table 1 
(OGD, FTL, and MA), specifically for the linear-quadratic case. 
The OGD update takes the form:
\beq
{\bf W}_{k+1}={\bf W}_k-\frac{1}{\eta_k}\nabla_{{\bf W}}\sum_{t=1}^{T_k}\ell({\bf W}_k,\textbf{z}_{kt},\ov_{kt}).
\eeq
The regret scales as $\mathcal{O}(\sqrt{K})$, so $\gamma_K$ is $\mathcal{O}(\frac{1}{\sqrt{K}})$ \cite{Kivinen1995}.  

Strong convexity on the loss $\sum_{t=1}^{T_k}\ell({\bf W}_k,\textbf{z}_{kt},\ov_{kt})$ for all $k$ will give a regret rate of $\mathcal{O}(\log K)$ \cite{Hazan2007}. 
This can be achieved by adding regularization on ${\bf W}$, which is typically done in practice. 
Alternatively, this condition will usually be satisfied when $T_k$ is greater than the number of parameters in ${\bf W}_k$, although this also depends on the data ${\bf z}_{kt}$ and $\ov_{kt}$.

The MA update is given by:
\beqs
{\bf W}_{k+1}&=&\lambda {\bf W_k}+(1-\lambda){\bf W}^\ast_k\\
{\bf W}^\ast_k&=&\arg\min \sum_{t=1}^{T_k}\ell({\bf W}_k,\textbf{z}_{kt},\ov_{kt})) 
\eeqs
for $\lambda\in[0,1]$.  As discussed in the main text, this algorithm suffers from regret that is at least $O(K)$, so it is not a no-regret algorithm.  

The FTL update is given by:
\beqs
&\pi^{(k)}=\arg\min_{\pi\in\Pi} \sum_{k'=1}^{k-1}\sum_{t=1}^{T_k}\ell(\pi,\textbf{z}_{k't},\ov_{k't}),\\
\Leftrightarrow & {\bf W}=\arg\min_{\bf W}\sum_{k'=1}^{k-1}\sum_{t=1}^{T_{k'}} ||{\bf W} \textbf{z}_{k't}-{\bf o}_{k't}||^2 +g({\bf W}).
\eeqs
$g({\bf W})$ is a optional regularization penalty.
Typical regret analysis for FTL depends on the Lipschitz properties of the loss function (for a smooth function, this implies that the gradient is bounded).  We emphasize that any standard optimization technique can be used here, and the FTL strategy is not sensitive to step-sizes.  Restricting the parameter set to a ball such that $||{\bf W}||_2^2\leq B$ and assuming $||{\bf z}_{kt}||_2\leq 1 ~ \forall k,t$ and $||{\bf o}_{kt}||_2\leq 1 ~ \forall k,t$ yields $\mathcal{L}(W,\mathcal{D}^{(1:k)})$ =  $\sum_{k'=1}^{k-1} \sum_{t=1}^{T_{k'}}\ell({\bf W},\textbf{z}_{k't},\ov_{k't})$ as $B^2$-Lipschitz, which can be used to analyze a bounded least-squared problem \cite{Hazan2007}.  These conditions will be satisfied when using feasible data generated in the system with a regularized $g({\bf W})$.

To get a better regret bound for FTL, we must analyze it through the perspective of another approach, called \textit{Exponentially-weighted online optimization} (EWOO) \cite{Hazan2007}.  We will not talk about this method in general; however, for the specific case of the least squares loss function and $\ell_2$ regularization, the updates for EWOO are identical to the updates for FTRL (or FTL if the regularization is omitted).  We emphasize that these updates are not equivalent in general.  This is beneficial in our case because it derives a logarithmic regret bound $\mathcal{O}(\frac{\ell_{\max}}{\alpha}\log K)$, and $\alpha$ will be described below.  In this case, $\gamma_K$ will scale as $\mathcal{O}(\log(K)/K)$, which is an improvement over the $\mathcal{O}(1/\sqrt{K})$ rate.  Instead of being dependent on Lipschitz smoothness of the loss function, the constants are dependent on an alternative property called $\alpha \textit{-} exp \textit{-} concavity$, which is defined:
\beq
\forall {\bf W} \in \mathcal{P},\forall k \in 1, \dots, K: \nabla^2[\exp(\alpha \mathcal{L}(W,\mathcal{D}^{(1:k)}))]\preceq 0.
\eeq
This property depends on a non-empty convex set, $\mathcal{P}\subseteq \mathbb{R}^P$, which corresponds to the \textit{feasible} parameters of the decoder/policy.    In general, this is unconstrained; however, given certain properties of the dataset, the set $\mathcal{P}$ can be quite constrained.  We note that any strongly convex function has $\alpha$\textit{-}exp\textit{-}concavity, but that this is a \textit{weaker} property than strong convexity.  For the least squares problems, without this assumption the $\alpha$ constant can be arbitrarily bad.   We will discuss reasonable assumptions below, and mention how they restrict $\mathcal{P}$ and therefore the constant $\alpha$ as well.
 
 Since $\alpha$ affects the regret of the algorithm, we need to get a sense of the value of $\alpha$ in practice in order to truly assess the performance of this algorithm.  For our case, $\alpha$ can be simplified to $
\frac{1}{\alpha}=\max_{{\bf W}\in \mathcal P,k,t}||{\bf W} \textbf{z}_{kt}-{\bf o}_{kt}||_2^2
$.  Next, we will utilize a standard trick, where $\mathcal{P}$ is set and analyzed over realizable values that ${\bf W}$ can take \cite{Shalev-Shwartz2011}.  To get a simpler form of this analysis, we will make the significant, but reasonable, assumption that the worst parameter settings will be our initialization, which we set to {\bf 0} here for simplicity.  This assumption makes $\frac{1}{\alpha}$ scale as $\mathcal{O}(||{\bf o}||_2^2)$.  The same assumption will set $\ell_{\max}$ to $\mathcal{O}(||{\bf o}||_2^2)$.  Hence, we expect the total regret to scale as $\mathcal{O}(||{\bf o} ||_2^4\log K)$.  This gives a strong, but reasonable, dependency on the magnitude of the oracle movements.

\subsection*{S1 Fig}
\label{fig:S1_Fig}
{\bf Comparison of MSE and acquisition time} {This figure compares MSE and time to acquisition for the cursor task, and motivates the use of SSE in the figures in the main text. Left panel depicts MSE for cursor task (for same trials as SSE curves in Fig. \ref{fig:cursor}). Right panel depicts time to acquisition for the same set of trials.  While we might hope that MSE would give a complete indication of performance, this is not the case.  This is because the quality of the different algorithms are differentially reflected when considering trial duration.  Low MSE can be achieved multiple different ways -- essentially mapping to the bias-variance tradeoff.  In the trials considered here, the slow acquisition for the MA decoder arises from bias towards decoder outputs with smaller magnitude.}  

\begin{figure}[h!]
\begin{center}
  \subfloat{\includegraphics[width=.7\textwidth]{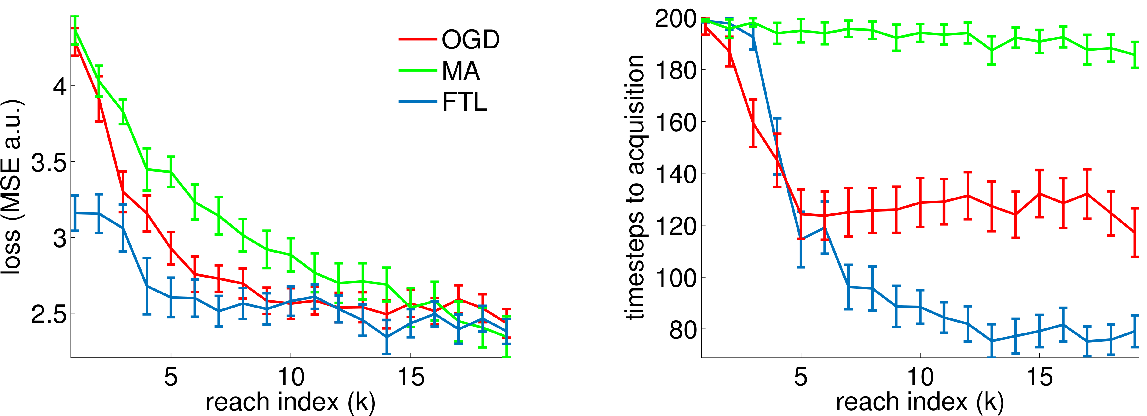}} 
  \end{center}    
  \captionsetup{width=.8\textwidth}
\captionsetup{font={footnotesize}}
\label{fig:im_supp_1}
\end{figure}

\subsection*{S1 Movie}
\label{mov:S1_mov}
{\bf Example cursor trials} Video of cursor task during learning via \textsc{DAgger}.  Blue dot corresponds to controlled cursor.  Green dot corresponds to target.  Green line from blue cursor points towards the target.  Red line from cursor corresponds to actual direction of motion.

\subsection*{S2 Movie}
\label{mov:S2_mov}
{\bf Example full-arm trials} Video of full arm task during learning via \textsc{DAgger}.  Arm is controlled to reach towards the wand.  Initial arm pose is reset between reaches.

\section*{Acknowledgments}
We'd like to thank the Pesaran Lab, especially Adam Weiss and Yan Wong, who provided assistance related to using an arm model of one of their monkeys.  
Chris Cueva helped with MATLAB scripts to interact with MuJoCo. 
Grace Lindsay contributed the illustration in Figure 1.

\section*{Funding}
This work was supported by ONR N00014-16-1-2176 and a Google Research Award to LP.  Simons Global Brain Research Awards 325171 and 325233 supported LP and JPC. JPC is supported by a Sloan Research Fellowship.  All authors receive support from the Grossman Center at Columbia University, and the Gatsby Charitable Trust.  The funders had no role in study design, data collection and analysis, decision to publish, or preparation of the manuscript.


{\bibliography{tex/references.bib}}

\end{document}